\documentclass[10pt,journal,compsoc]{IEEEtran}

\usepackage{color}
\usepackage[normalem]{ulem}

\ifCLASSOPTIONcompsoc
  \usepackage[nocompress]{cite}
\else
  \usepackage{cite}
\fi

\ifCLASSINFOpdf
   \usepackage[pdftex]{graphicx}
   \graphicspath{{../pdf/}{../jpeg/}{./figs/}{./figures}}
\else
\fi

\usepackage{graphicx}
\ifCLASSINFOpdf

\else

\fi

\usepackage[cmex10]{amsmath}
\usepackage{array}
\usepackage{multirow}
\usepackage{lineno}
\usepackage{url}
\usepackage{amsmath}

\def\0{{\bf 0}}
\def\1{{\bf 1}}

\usepackage{amsmath}
\usepackage{amssymb}
\usepackage{multirow}
\usepackage{array}
\usepackage{booktabs}
\usepackage{multirow}
\usepackage{bm}
\usepackage{graphicx}
\usepackage{epstopdf}
\usepackage{subfigure}
\usepackage{makecell}
\usepackage[pagebackref=false,breaklinks=true,letterpaper=true,colorlinks,bookmarks=false]{hyperref}
\usepackage{cite}
\usepackage{multirow}
\usepackage{color}
\usepackage{CJK}
\usepackage{soul}
\usepackage[numbers,sort&compress]{natbib}
\usepackage{hyperref}

\usepackage[colorinlistoftodos]{todonotes}

\begin{document}

\title{Learning from Heterogeneous Structural MRI via Collaborative Domain Adaptation for Late-Life Depression Assessment}

\author{Yuzhen Gao, 
Qianqian~Wang,
Yongheng Sun, 
Cui Wang,
Yongquan Liang, 
Mingxia~Liu,~\IEEEmembership{Senior Member,~IEEE}

\IEEEcompsocitemizethanks{
\IEEEcompsocthanksitem Y.~Gao, Q.~Wang, Y.~Sun, C. Wang, and M.~Liu are with the Department of Radiology and Biomedical Research Imaging Center (BRIC), University of North Carolina at Chapel Hill, Chapel Hill, NC 27599, USA. 
Y.~Liang is with the School of Computer Science and Engineering, Shandong University of Science and Technology, Qingdao 266590, China. 
\IEEEcompsocthanksitem 
Corresponding author: M.~Liu (Email: mingxia\_liu@med.unc.edu). 
\protect\\
}
}
{}

\IEEEtitleabstractindextext{%
\begin{abstract}
Accurate identification of late-life depression (LLD) using structural brain MRI is essential for monitoring disease progression and facilitating timely intervention. However, existing learning-based approaches for LLD detection are often constrained by limited sample sizes (e.g., tens), which poses significant challenges for reliable model training and generalization. Although incorporating auxiliary datasets can expand the training set, substantial domain heterogeneity, such as differences in imaging protocols, scanner hardware, and population demographics, often undermines cross-domain transferability.
To address this issue, we propose a Collaborative Domain Adaptation (CDA) framework for LLD detection using T1-weighted MRIs. 
The CDA leverages a Vision Transformer (ViT) to capture global anatomical context and a Convolutional Neural Network (CNN) to extract local structural features, with each branch comprising an encoder and a classifier. 
The CDA framework consists of three stages: (a) \emph{supervised training} on labeled source data, (b) \emph{self-supervised target feature adaptation} and (c) \emph{collaborative training} on unlabeled target data. 
We first train ViT and CNN on source data, followed by self-supervised target feature adaptation by minimizing the discrepancy between classifier outputs from two branches to make the categorical boundary clearer. 
The collaborative training stage employs pseudo-labeled and augmented target-domain MRIs, enforcing prediction consistency under strong and weak augmentation to enhance domain robustness and generalization. 
Extensive experiments conducted on 
multi-site T1-weighted MRI data 
demonstrate that the CDA consistently outperforms state-of-the-art unsupervised domain adaptation methods.
\end{abstract}

}

\maketitle

\IEEEdisplaynontitleabstractindextext

\IEEEpeerreviewmaketitle

\IEEEraisesectionheading{\section{Introduction}\label{S1}}

\IEEEPARstart{L}{ate-life} depression (LLD) is a prevalent and debilitating neuropsychiatric disorder that affects a substantial proportion of the aging population. It poses significant challenges to cognitive health, quality of life, and overall mortality rates ~\cite{blazer2003depression, alexopoulos2005vascular}. 
Early and accurate identification of LLD is essential to facilitate timely intervention and mitigate the risk of progression to more severe cognitive impairment. 
Neuroimaging, particularly structural magnetic resonance imaging (MRI), has emerged as a valuable solution for exploring the neurobiological basis of LLD. 
Prior MRI-based studies have reported that LLD is associated with widespread structural abnormalities, particularly in regions such as the prefrontal cortex, hippocampus, and major white matter tracts~\cite{krishnan1997neuroanatomical,sawyer2012depression, mortimer2013neuroimaging}.

Learning-based approaches, including both traditional and deep models, have demonstrated potential in MRI biomarker identification and  automated classification of LLD~\cite{herrmann2008white,sexton2013systematic}. 
Unfortunately, these models rely on large-scale training data to achieve robust performance~\cite{jonsson2019brain,knoll2020deep,maartensson2020reliability,wen2020convolutional,eitel2021promises}, whereas LLD-related MRI datasets are typically limited in size. 
While aggregating multiple neuroimaging datasets can increase sample size and improve model generalization, this approach often introduces domain shift~\cite{ganin2015unsupervised,maartensson2020reliability, sarafraz2024domain}, which refers to statistical differences caused by heterogeneous scanner types, imaging protocols, and subject populations.

Domain adaptation techniques have attracted increasing attention in these data-constrained medical imaging scenarios, aiming to reduce the discrepancy between source and target distributions~\cite{guan2023domainatm}.
Classical methods include adversarial training~\cite{ganin2015unsupervised}, maximum mean discrepancy (MMD) minimization~\cite{long2015learning}, and self-supervised consistency regularization~\cite{french2018self, berthelot2019mixmatch}. 
These techniques have demonstrated promising results in medical imaging tasks such as brain tumor segmentation~\cite{dou2018unsupervised}, Alzheimer's disease classification~\cite{pan2021disease,agarwal2021transfer}, and schizophrenia prediction~\cite{liu2023transfer}.
However, current domain adaptation approaches face two key limitations. 
First, they often treat the model as a fixed, uniform system 
and focus solely on aligning feature distributions across domains, often overlooking the complementary nature of different model architectures or representations~\cite{pei2018multi, xu2020adversarial}. 
Second, they often struggle to achieve robust performance when the target domain has extremely limited data, a common issue in the context of LLD~\cite{zhao2021domain, perone2019unsupervised}.

\begin{figure*}[!t]
\setlength{\belowdisplayskip}{-1pt}
\setlength{\abovedisplayskip}{-1pt}
\setlength{\abovecaptionskip}{-1pt}
\setlength{\belowcaptionskip}{-1pt}
\centering
\includegraphics[width=0.98\textwidth]{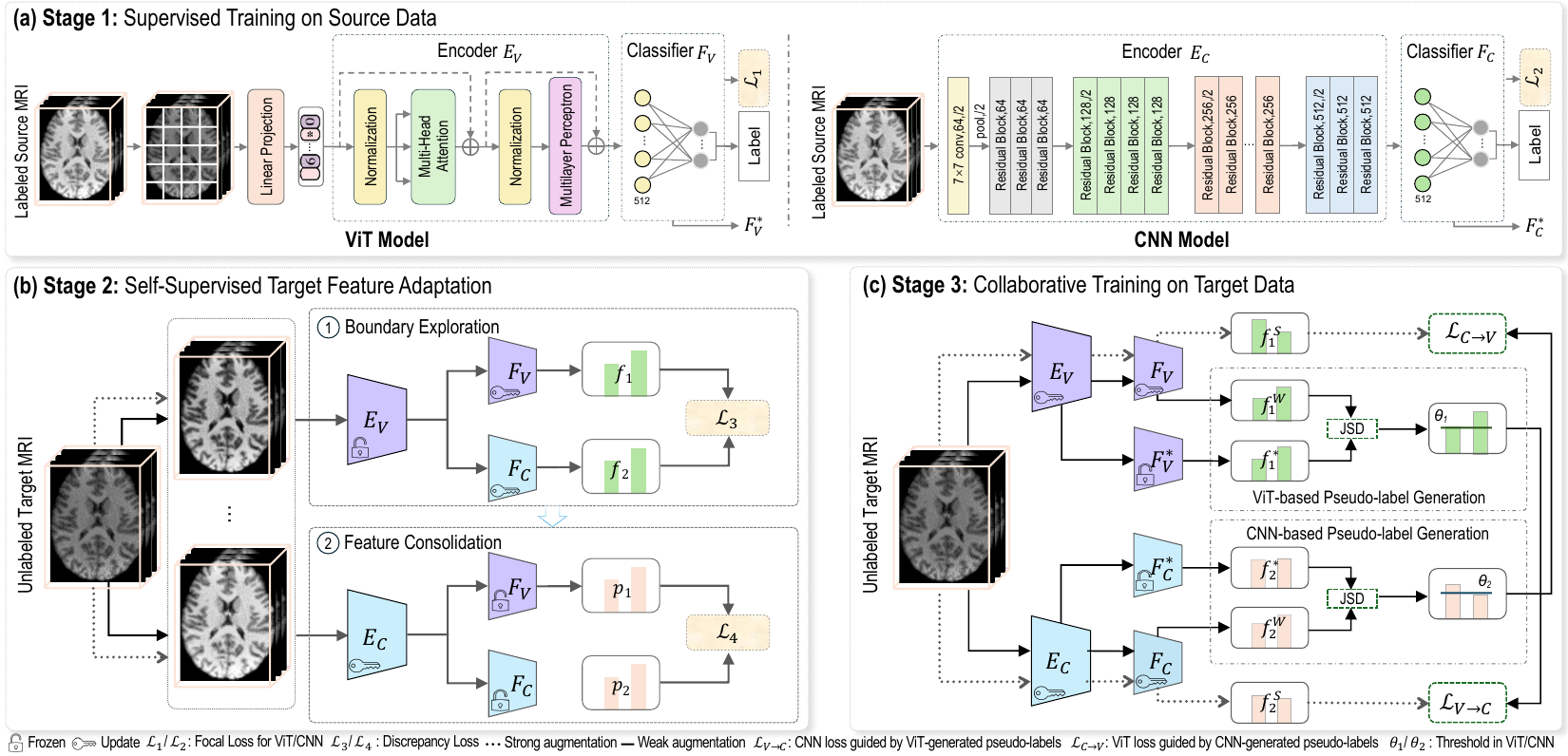}
\caption{Illustration of collaborative domain adaptation (CDA), consisting of (a) supervised training on labeled source data, (b) self-supervised target feature adaptation between ViT and CNN branches to refine decision boundaries, and (c) collaborative training using pseudo-labeled and augmented target samples to enhance prediction consistency and domain generalization. The ViT branch consists of an encoder $E_V$ and a classifier $F_V$, while the CNN branch comprises an encoder $E_C$ and a classifier $F_C$. JSD: Jensen-Shannon divergence.}
\label{CDA}
\end{figure*}

To this end, we propose a novel Collaborative Domain Adaptation (\textbf{CDA}) framework for MRI-based identification of LLD. 
Our framework integrates a Vision Transformer (ViT){~\cite{dosovitskiy2020image}} 
to model global anatomical features and a Convolutional Neural Network (CNN)~\cite{he2016deep} to capture local structural patterns from MRI. 
The complementary nature of these architectures enables richer representation learning. 
As illustrated in Fig.~\ref{CDA}, the CDA framework comprises three stages: (a) \emph{supervised training} of ViT and CNN on labeled source data, (b) \emph{self-supervised target feature adaptation}, and (c) \emph{collaborative training} on unlabeled target data. 
The first stage uses supervised learning on the source domain to ensure that ViT and CNN branches have basic discriminative capabilities. 
During the self-supervised target adaptation stage,
the two models are fine-tuned through feature alignment to minimize the discrepancy between the classifier outputs of the two branches, thereby refining the decision boundaries. 
The subsequent collaborative training stage utilizes pseudo-labeled and augmented target-domain MRIs to enforce prediction consistency under both strong and weak perturbations, enhancing domain robustness and generalization.
This collaborative strategy effectively addresses both the data scarcity and domain shift problems by leveraging source knowledge while promoting generalization to the target domain. 
Experiments on multi-site T1-weighted MRI data related to late-life depression demonstrate that CDA significantly outperforms several state-of-the-art domain adaptation baselines across multiple evaluation metrics. 
The source code has been released to the public through  GitHub\footnote{https://github.com/yzgao2017/CDA}.

The contributions of this work are summarized below.
\begin{itemize}
\vspace{-2pt}
    \item We propose a Collaborative Domain Adaptation (CDA) framework for automated LLD identification using structural MRI. The framework features a dual-branch architecture that integrates a Vision Transformer (ViT) and a Convolutional Neural Network (CNN), one of the first efforts to combine these complementary models in a domain adaptation setting. This design enables the model to capture both global anatomical context and local structural details effectively. 
    \item We introduce a novel three-stage training strategy comprising: (1) supervised learning on labeled source data, (2) self-supervised target feature adaptation for decision boundary refinement, and (3) collaborative training to improve robustness in unlabeled target domains. This paradigm facilitates effective source-to-target knowledge transfer while maintaining strong predictive performance. 
    \item The collaborative training stage employs a unique mechanism that leverages pseudo-labels generated by both ViT and CNN branches, along with augmented target samples. This enhances model robustness and adaptation performance, particularly in data-limited target domains. 
    \item Extensive experiments on real multi-site structural MRI datasets demonstrate that CDA outperforms existing state-of-the-art domain adaptation methods in terms of classification accuracy and generalization ability, highlighting its practical value in cross-site neuroimaging applications.
\end{itemize}

The remainder of this paper is organized as follows. 
Section~\ref{S2} reviews the relevant studies. 
Section~\ref{S3} introduces materials and data pre-processing. 
Section~\ref{S4} provides a comprehensive description of the proposed method. 
Section~\ref{S5} introduces experimental setup and results. 
Section~\ref{S6} presents an ablation study, analyzes the influences of key CDA components, 
and discusses the limitations of the current work. 
This paper is concluded in Section~\ref{S7}.

\section{Related Work}
\label{S2}
 
\subsection{Late-Life Depression Progression Analysis}

Late-life depression (LLD) is a prevalent but often underdiagnosed neuropsychiatric disorder affecting a significant proportion of the elderly population. 
Growing evidence links LLD to accelerated cognitive decline and an increased risk of progression to dementia~\cite{butters2008pathways,mast2008vascular}. 
To better understand its neurobiological underpinnings, structural magnetic resonance imaging (MRI) has been widely used to investigate neuroanatomical changes associated with LLD. 
Consistent alterations have been reported in brain regions including the prefrontal cortex, hippocampus, amygdala, and anterior cingulate cortex~\cite{bora2012gray,lai2013gray}. 
Recent studies have increasingly focused on characterizing the progression of LLD and its potential biomarkers using structural MRI~\cite{alexopoulos2009serotonin,gunning2009anterior,koenig2014cognition}. 
However, many of these investigations rely on small, homogeneous cohorts or data collected at a single site, which limits statistical power and hinders the generalizability of the findings~\cite{benjamin2011structural, taylor2013vascular, kanellopoulos2011hippocampal}.

Given the scarcity of large-scale, multi-site datasets for LLD, recent efforts have focused on pooling data from heterogeneous sources or conducting cross-cohort analyses to increase sample size and diversity~\cite{wen2021multidimensional,tan2023disrupted}. 
While such strategies can enhance the representativeness of findings, they also introduce substantial \emph{domain shifts} due to differences in scanner hardware, imaging protocols, clinical labeling standards, and population characteristics. 
These variations significantly affect model performance and reproducibility, particularly in neuroimaging-based diagnostic tasks, where models trained on one dataset often struggle to generalize to others without proper domain adaptation strategies~\cite{ghafoorian2017transfer, wachinger2021detect, wolleb2022learn}. 
These challenges highlight the urgent need for robust, domain-adaptive learning approaches capable of addressing cross-site heterogeneity in 3D structural MRI for reliable LLD diagnosis.

\subsection{Domain Adaptation for MRI Analysis}

Domain adaptation (DA) has become an essential research direction for mitigating domain shift between training and testing distributions, particularly in medical imaging, where variability across scanners, institutions, and acquisition protocols is inevitable. 
Among various DA paradigms, unsupervised domain adaptation (UDA) has gained increasing attention for its potential to improve generalization in real-world clinical deployments~\cite{tzeng2017adversarial,jiang2022disentangled}, which assumes access to labeled source data and unlabeled target data.

Common strategies in UDA include adversarial learning~\cite{tzeng2017adversarial,kamnitsas2017unsupervised}, distribution alignment via maximum mean discrepancy (MMD)~\cite{long2015learning}, and entropy minimization~\cite{vu2019advent}. 
In neuroimaging, these techniques have been applied to tasks such as tumor segmentation~\cite{kamnitsas2017unsupervised} and Alzheimer's staging~\cite{guan2021multi}. 
Most existing DA models rely on CNN-based architectures due to their local receptive field and inductive bias towards translational invariance. 
However, CNNs often fall short in capturing long-range dependencies and holistic anatomical patterns, especially in high-resolution 3D MRI volumes.

In contrast, Vision Transformers (ViTs) have emerged as powerful alternatives, offering the ability to model global anatomical context through self-attention mechanisms.  
Recent models such as TransUNet~\cite{chen2021transunet} and UNETR~\cite{hatamizadeh2022unetr} have demonstrated state-of-the-art performance in medical image segmentation and classification. 
Nevertheless, most existing UDA approaches rely exclusively on either CNN or ViT backbones~\cite{liu2021swin, huang2020self}, failing to fully exploit the complementary strengths of both. 
Our work addresses this gap by introducing a hybrid domain adaptation framework that integrates ViT and CNN in a collaborative dual-branch architecture. This design leverages ViT's ability to capture global context and CNN’s strength in modeling fine-grained local features, resulting in a more comprehensive feature representation.

\subsection{Collaborative Learning in Domain Adaptation}
Collaborative learning strategies have been proposed to improve DA model robustness under limited target supervision. 
In particular,
co-training~\cite{blum1998combining} and mutual pseudo-labeling approaches~\cite{shu2018dirt, you2019universal, tang2020unsupervised} iteratively refine predictions across multiple views or learners, enhancing model stability in UDA settings. 
 These techniques are especially valuable in medical imaging tasks, where obtaining ground-truth labels in the target domain is costly or infeasible.

More recently, hybrid ViT-CNN architectures have been explored to capture complementary structural priors in complex imaging data~\cite{xu2023dual, dong2025multi}. 
A recent study~\cite{ngo2024learning} introduced an explicitly class-specific boundaries (ECB) learning framework, in which the ViT branch maximizes classifier discrepancy to explore class boundaries, while the CNN branch minimizes this discrepancy to cluster features within each class. 
This asymmetric design enhances decision margin clarity and enables more accurate pseudo-label propagation. 
Inspired by ECB, our method extends this paradigm by integrating co-training with supervised classifier feedback. 
The ViT branch maximizes inter-class separation, while the CNN branch refines intra-class consistency via class-conditional alignment. 
This dual-branch collaboration enhances prediction reliability and mitigates domain bias in LLD classification, effectively leveraging the complementary strengths of ViT and CNN in domain adaptation.

\begin{table}[!t]
\renewcommand\arraystretch{1}
\centering
\scriptsize
\caption{Demographic information and category labels of the subjects in cognitive impairment-related studies from different data domains in the article. The values are denoted as ``mean $\pm$ standard deviation''. F/M: Female/Male, MMSE: Mini-Mental State Examination.}

\resizebox{0.9\linewidth}{!}{  
\begin{tabular}{llllll}   
\toprule
Domain & Dataset & Category & Gender (F/M) & Age & MMSE \\
\midrule
\multirow{4}{*}{Target} & \multirow{4}{*}{NBOLD} 
& CN-N  & 27/7 & 71.6$\pm$7.1 & 29.1$\pm$1.5 \\
& & CN-D  & 50/16 & 74.9$\pm$4.3 & 29.2$\pm$1.2 \\
& & CIND  & 6/3 & 74.9$\pm$4.3 & 29.2$\pm$1.2 \\
& & AD    & 3/5 & 76.5$\pm$7.5 & 28.5$\pm$1.7 \\
\midrule 
\multirow{4}{*}{Source} & \multirow{4}{*}{NCODE} 
& CN-N  & 33/23 & 67.5$\pm$5.0  & 29.1$\pm$1.0 \\
& & CN-D  & 69/41 & 66.1$\pm$5.7 & 28.5$\pm$1.4 \\
& & CIND  & 5/5 & 71.9$\pm$6.3 & 27.4$\pm$2.2 \\
& & AD    & 5/3 & 72.0$\pm$5.6 & 24.8$\pm$5.4 \\
\bottomrule
\end{tabular}
}
\label{tabel2}
\end{table}

\section{Materials}
\label{S3}


\subsection{Studied Subjects}
This study uses multi-site T1-weighted MRI data related to late-life depression from two imaging cohorts: 
1) the Neurobiology of Late-life Depression (NBOLD) database~\cite{steffens2017negative}, and 2) the Neurocognitive Outcomes of Depression in the Elderly (NCODE) study~\cite{steffens2004ncode}.  
The demographic and clinical characteristics of participants are reported in Table~\ref {tabel2}.

\textbf{NBOLD}. The NBOLD study collected T1-weighted MRI scans between 2013 and 2017 at the Olin Neuropsychiatric Research Center. 
The scans were acquired using a Skyra 3T MRI scanner with MPRAGE protocol (TR/TE = 2200/2.88 ms, voxel size = 0.8$\times$0.8$\times$0.8 mm). 
All participants were older adults assessed for late-life depression (LLD) and followed up for cognitive outcomes. 
Participants were enrolled at baseline with cognitive normality and monitored for five years. 
Diagnoses were assigned based on five-year follow-up outcomes and clinical consensus, resulting in four categories: 1) never-depressed, cognitively normal (CN-N); 2) depressed, cognitively normal (CN-D); 3) cognitively impaired but without dementia  (CIND); and 4) Alzheimer's disease (AD), as presented in Table~\ref{table_Category}. 
All diagnoses were confirmed by experienced geriatric psychiatrists using standardized protocols and clinical scales. 
Considering the limited sample size of CIND and AD, we combined these two groups into the cognitive impairment (CI) category. 
This dataset was used as the \emph{target domain} in this work, including 34 CN-N subjects, 66 CN-D subjects, and 17 CI subjects (CIND: 9, AD: 8).

\textbf{NCODE}. The NCODE includes 3T T1-weighted MRI scans collected between 1995 and 2013 using a Siemens Trio MRI scanner. 
These scans were acquired using 3D TURBOFLASH sequence (TR/TE = 22/7 ms, voxel size = 1$\times$1$\times$1 mm). These scans provide detailed brain structural information relevant for both depression and cognitive decline analysis.  
As in NBOLD, all participants were cognitively normal at baseline and were followed for five years. 
Their diagnostic outcomes at the five-year follow-up were used for classification. 
Subjects were categorized into three diagnostic groups: CN-N, CN-D, and CI (with CIND and AD). 
This dataset was used as the \emph{source domain} in this work, consisting of 56 CN-N subjects, 110 CN-D subjects, and 18 CI subjects (CIND: 10, AD: 8).

\begin{table}[!t]
\renewcommand\arraystretch{1}
\centering
\scriptsize
\caption{Diagnostic category and description of the studied participants.
}
\resizebox{0.9\linewidth}{!}{
\begin{tabular}{lp{6cm}}  
\toprule
\textbf{Category} & \textbf{Description} \\
\midrule
CN-N & Cognitively Normal \\
CN-D & Depressed, Cognitively Normal \\
CIND & Cognitive Impairment, No Dementia;\\
     & Cognitive Impairment, due to Vascular Disease \\
AD   & Probable AD, Possible AD, Subsyndromal AD;\\
     & Dementia of undetermined etiology \\
\bottomrule
\end{tabular}
}
\label{table_Category}
\end{table}

\subsection{Data Pre-Processing}
All MRI scans underwent a standard preprocessing pipeline to ensure consistency across datasets. First, N4 bias field correction was applied to address intensity inhomogeneities~\cite{tustison2010n4itk}. 
Then, images were linearly and non-linearly registered to the MNI space using the Advanced Normalization Tools (ANTs)~\cite{avants2011ants}. 
Skull stripping was performed using a robust brain extraction method. 
All images were resampled to 1\,mm isotropic resolution and normalized in shape to the Automated Anatomical Labeling 3 (AAL3) atlas~\cite{rolls2020aal3}, which includes 170 regions of interest (ROIs). 
Structural brain features such as cortical thickness, surface area, and gray matter volume were extracted using FreeSurfer~\cite{fischl2012freesurfer}. Histogram matching and z-score normalization were performed using TorchIO~\cite{perez-garcia2021torchio} to standardize intensity distributions across sites. 
The final images were standardized to 181$\times$217$\times$181 voxels in the Montreal Neurological Institute (MNI) space, ensuring spatial correspondence across all datasets.

\section{Methodology}
\label{S4}
\textbf{Problem Formulation}.  
The goal is to identify patients with LLD from cognitively normal subjects in a target domain with  limited data, by leveraging knowledge from relatively large-scale source data. 
We have a {labeled source domain} 
{\small{$D_S$$=$$ \{(x^s_i, y^s_i)\}^{N_s}_{i=1} $}},  
where each 3D MRI {\small{\( x^s_i \in \mathbb{R}^{H \times W \times D} \)}} corresponds to a ground-truth category label {\small{\( y^s_i 
\)}}, and 
an {unlabeled target domain}      {\small{\( D_T = \{x^t_i\}^{N_t}_{i=1} \)}},
where $N_s$ is the number of source samples and $N_t$ is the number of target samples. 
As illustrated in Fig.~\ref{CDA}, the proposed CDA has two branches: 1) ViT for global MRI feature learning, and 2) CNN for local MRI feature learning. 
This dual-branch design is motivated by the fact that ViT can effectively capture long-range dependencies and model comprehensive global representations, while CNN is good at capturing local spatial hierarchies for robust feature learning. 
To facilitate  knowledge transfer from source to target domain, we develop three components in CDA, including 1) \emph{supervised training} on labeled source data,  2) \emph{self-supervised target feature adaptation}, and 3) \emph{collaborative training} of ViT and CNN on unlabeled target data. 
To improve cross-domain generalization, we employ data augmentation in CDA, 
applying weak and strong augmentations~\cite{cardoso2022monai} to enhance their feature consistency.

\subsection{Supervised Training on Labeled Source Data}

As shown in Fig.~\ref{CDA}~(a), each input MRI image is simultaneously processed by two distinct branches: a ViT and a CNN, each comprising an encoder and a classifier. 
The ViT branch first divides the 3D MRI into non-overlapping patches, which are then embedded into tokens and preprocessed by an encoder {{$E_V$}} composed of multiple self-attention layers. 
The resulting global feature representation is fed into a classifier {{$F_V$}} for category prediction. 
In parallel, the CNN branch extracts hierarchical local features through an encoder {{$E_C$}} consisting of stacked convolutional layers and residual blocks. 
These features are then input to the CNN classifier {{$F_C$}}, yielding the corresponding class logits.
{To ensure effective feature learning, both branches are independently trained on labeled source domain samples.}
Given the potential class imbalance in the source data, we employ the focal loss~\cite{lin2017focal} to increase the model's sensitivity to hard or minority class examples. 
The optimization objectives for the two branches are defined as:
\begin{equation}
\small
\begin{split}
\mathcal{L}_{1} = \frac{1}{N_s} \sum\nolimits_{i=1}^{N_s} 
\mathcal{FL}(y_i^s, \sigma(F_V(E_V(x_i^s)))),
\\
\mathcal{L}_{2} = \frac{1}{N_s} \sum\nolimits_{i=1}^{N_s} 
\mathcal{FL}(y_i^s, \sigma(F_C(E_C(x_i^s)))), 
\end{split}
\end{equation}
where {{$\mathcal{FL}(\cdot)$}} is the focal loss, 
{{$\sigma(F_V(E_V(x_i^s)))$}} and 
{{$\sigma(F_C(E_C(x_i^s))))$}} denote the predicted probability scores from the ViT and CNN branches, respectively, and $y_i^s$ is the ground-truth category label of the source sample $x_i^s$. 
Here, the softmax function
{{$\sigma(\cdot)$}} is used to convert logits into normalized class probabilities. 
The generated classifiers (i.e., {{$F_V^*$}} and {{$F_C^*$}}) trained on labeled source data will be utilized in the third stage of the CDA framework.

\subsection{Self-Supervised Target Feature Adaptation}

In this work, we aim to bridge the domain gap between the source and target distributions by aligning their feature representations in a self-supervised manner. 
Specifically, we leverage the complementary properties of ViT and CNN branches: the ViT encoder defines sharper and more discriminative class boundaries, while the CNN encoder adjusts its representations to align with those boundaries. 
This module comprises two sequential phases, as illustrated in Fig.~\ref{CDA}~(b): (1) \emph{Boundary Exploration}, and (2) \emph{Feature Consolidation}. 

\emph{\textbf{1) Boundary Exploration}}:
In this phase, the ViT encoder {{\(E_V\)}} is frozen, and the classifiers {{\(F_V\)}} and {{\(F_C\)}} are fine-tuned to maximize their predictive divergence on unlabeled target samples. 
This encourages the two classifiers to expand their decision boundaries, facilitating the identification of target features that deviate from the support of source distribution. 
To quantify disagreement, we define a discrepancy loss between class probability outputs {{$\mathbf{a}$}} and {{$\mathbf{b}$}} of the two classifiers as:
\begin{equation}
\mathcal{DL}(\mathbf{a}, \mathbf{b}) = \frac{1}{K} \sum\nolimits_{k=1}^{K} |a_k - b_k|,
\label{eq:D_ab}
\end{equation}
where $K$ is the number of classes, and $a_k$ and $b_k$ denote the predicted probabilities from $F_V$ and $F_C$ for the $k$-th class. 
Specifically, we fix the encoder {{\(E_V\)}} (trained in Stage 1) 
and update the classifiers {{\(F_V\)}} and {{\(F_C\)}} to maximize their prediction difference on target data. 
The objective function is defined as:
\begin{equation}
\small
\begin{split}
\mathcal{L}_{3}= 
 &-\frac{1}{N_{t}}\sum\nolimits_{i=1}^{N_{t}}
   \mathcal{DL} (f_1, f_2)       
       \\
 &+\frac{1}{N_{s}}\sum\nolimits_{i=1}^{N_{s}}
   \mathcal{FL} \bigl(y_{i}^{s},\,
            \sigma\!\bigl(F_{V}(E_{V}(x_{i}^{s}))\bigr)
   \bigr) \\
 &+\frac{1}{N_{s}}\sum\nolimits_{i=1}^{N_{s}}
   \mathcal{FL} \bigl(y_{i}^{s},\,
            \sigma\!\bigl(F_{C}(E_{C}(x_{i}^{s}))\bigr)
   \bigr), 
\label{eq_L3}
\end{split}
\end{equation}
where $f_1=\sigma(F_V(E_V(x_i^t)))$ and $f_2=\sigma(F_C(E_V(x_i^t)))$ are the softmax-normalized outputs of the classifiers for the unlabeled target sample $x_i^t$. 
Minimizing $\mathcal{L}_3$ encourages the classifiers to disagree on ambiguous target samples, thereby delineating class boundaries. 
We also input labeled source data to ViT and CNN for supervised training by minimizing the focal losses (i.e., the last two focal loss terms in Eq.~\eqref{eq_L3}),  
where their encoders (i.e., $E_V$ and $E_C$) are kept frozen and only the classifiers (i.e., $F_V$ and $F_C$) are updated. 
This enables the two classifiers to accurately distinguish between different categories without being misled by challenging target samples~\cite{saito2018maximum}. 
During training, we maintain a balanced number of source and target samples in each mini-batch to facilitate effective domain adaptation.

\emph{\textbf{2) Feature Consolidation}}:
To improve feature consistency across encoders, we refine the CNN encoder $E_C$ to better align its extracted features with the class boundaries defined by the ViT encoder $E_V$, particularly in regions of high uncertainty identified via classifier divergence.  
In this phase, the classifiers $F_V$ and $F_C$ are frozen, and $E_C$ is optimized to minimize the discrepancy between their predictions for target data:
\begin{equation}
\mathcal{L}_{4} = \frac{1}{N_{t}} \sum\nolimits_{i=1}^{N_{t}} \mathcal{DL}(p_1, p_2),
\end{equation}
where $p_1 = \sigma(F_V(E_C(x_i^t)))$ and $p_2 = \sigma(F_C(E_C(x_i^t)))$ denote the outputs of the two classifiers when applied to features extracted by the CNN encoder $E_C$. 
This procedure encourages the CNN to produce representations that are consistent with the previously established decision boundaries, thereby consolidating the feature space. 
Aligning features in this manner promotes classification robustness and generalization, particularly in the absence of labeled target samples.

\subsection{Collaborative Training on Unlabeled Target Data}

This phase aims to mitigate the discrepancy between the two network branches by enabling mutual reinforcement, leveraging their complementary strengths to generate reliable pseudo-labels for the unlabeled target samples. 
Notably, this approach enhances the CNN branch by exploiting the ViT’s superior capacity for modeling complex global patterns. 
To facilitate effective cross-branch knowledge transfer, pseudo-labels are generated by the ViT from weakly augmented target samples ($x^{t,W}$) and then employed to supervise the CNN branch on strongly augmented counterparts ($x^{t,S}$). Similarly, the CNN also guides the ViT under the same paradigm, as illustrated in Fig.~\ref{CDA}~(c). Strong augmentations are realized through random large-scale affine transformations and elastic deformations, while weak augmentations include random flipping and mild affine adjustments~\cite{cardoso2022monai}. To enhance the fidelity of pseudo-labels, a Jensen-Shannon divergence (JSD)-based optimization mechanism~\cite{menendez1997jensen} is incorporated.

\emph{\textbf{1) Pseudo-Label Generation:}} For each unlabeled target instance, feature extraction is performed by the encoders $E_V$ and $E_C$, followed by classification via $F_V$ and $F_C$ to obtain pseudo-label predictions. 
The preliminary classifiers $F_V^*$ and $F_C^*$, trained during the supervised stage, provide a reference for consistency evaluation on similarly augmented samples. The classification outputs on a set of weakly augmented inputs $x^{t,W}$ are denoted as: 
(a) $q_1^* = F_V^*(E_V(x^{t,W}))$ and $q_1 = F_V(E_V(x^{t,W}))$ for the ViT branch, where $F_V$ is the current classifier; 
(b) $q_2^* = F_C^*(E_C(x^{t,W}))$ and $q_2 = F_C(E_C(x^{t,W}))$ for the CNN branch, where $F_C$ is the current classifier. 
The Jensen-Shannon divergence (JSD) quantifies the consistency between preliminary and current classifier outputs within each branch, defined as:
\begin{equation}
JSD(P_1, P_2) = \frac{1}{2} D_{KL}(P_1 \| M) + \frac{1}{2} D_{KL}(P_2 \| M),
\end{equation}
where $M = \frac{1}{2}(P_1 + P_2)$ denotes the mixture distribution and $D_{KL}$ represents the Kullback-Leibler divergence~\cite{kullback1951information}. Consistency measures $JSD(q_1^*, q_1)$ and $JSD(q_2^*, q_2)$ are computed for ViT and CNN respectively. 
When the divergence falls below a predefined threshold $\tau = 0.1$, predictions are deemed sufficiently consistent, thus qualifying the corresponding samples for high-confidence pseudo-label generation. 
For such target samples within a batch, pseudo-labels are computed as:
\begin{equation}
\hat{y}^t_V = \frac{1}{2}(f_1^* + f_1^W), \quad
\hat{y}^t_C = \frac{1}{2}(f_2^* + f_2^W),
\end{equation}
where $f_1^* = \sigma(q_1^*)$, $f_1^W = \sigma(q_1)$, $f_2^* = \sigma(q_2^*)$, and $f_2^W = \sigma(q_2)$, and $\sigma(\cdot)$ denotes the softmax function.

\emph{\textbf{2) Collaborative Training:}}
To foster cross-branch learning, we employ pseudo-labels $\hat{y}_V^t$ generated by ViT for weakly augmented samples to supervise the CNN branch on strongly augmented inputs $x^{t,S}$ by minimizing:
\begin{equation}
\mathcal{L}_{V \to C} = CE\big(F_C(E_C(x^{t,S})), \hat{y}_V^t\big) \cdot \mathbf{1}\big(\max(\hat{y}_V^t) > \theta_1\big),
\label{eq_lossVITtoCNN}
\end{equation}
where $\mathbf{1}(\cdot)$ is the indicator function selecting only those samples with confidence exceeding threshold $\theta_1$, and $CE(\cdot)$ denotes the cross-entropy loss. 
Similarly, the CNN-generated pseudo-labels $\hat{y}_C^t$ supervise the ViT branch through:
\begin{equation}
\mathcal{L}_{C \to V} = CE\big(F_V(E_V(x^{t,S})), \hat{y}_C^t\big) \cdot \mathbf{1}\big(\max(\hat{y}_C^t) > \theta_2\big),
\label{eq_lossCNNtoVIT}
\end{equation}
where $\theta_2$ serves as the confidence threshold for CNN. 
These bidirectional supervisory signals promote effective knowledge exchange between the ViT and CNN branches, enhancing their respective capabilities in capturing global contextual information and local spatial hierarchies, encouraging robust MRI feature learning. 
During inference, we use the trained CNN model for prediction, with more discussions presented in Section~\ref{S6_EffectInference}. 

\subsection{Implementation}
In this work, we use a lightweight ViT architecture~\cite{touvron2021training} and a ResNet-34 CNN model from the MONAI framework~\cite{cardoso2022monai} in the proposed CDA framework. 
To leverage large-scale auxiliary  data, the ViT encoder $E_V$ is pretrained using a self-supervised Masked Autoencoder (MAE) approach on unlabeled brain MRI datasets (i.e., IXI, OASIS-3, and BRATS)~\cite{kunanbayev2024training}. 
The CNN encoder $E_C$ is pretrained on 9,544 T1-weighted MRIs from ADNI~\cite{jack2008alzheimer} via an autoencoder-based unsupervised image reconstruction approach~\cite{zhang2025brain}. 
Each classifier ($F_V$ for ViT and $F_C$ for CNN) consists of two fully connected layers 
followed by a softmax activation function. 
Model optimization employs Stochastic Gradient Descent (SGD)~\cite{bottou2012stochastic} with momentum 0.9 to accelerate convergence and escape local minima. 
A weight decay of 0.0005 provides $l_2$ regularization to improve generalization. 
Learning rates for both encoders are set to $1 \times 10^{-4}$ and $5 \times 10^{-4}$, respectively, appropriate for fine-tuning pretrained models. 
The batch size is set to 4, balancing GPU memory constraints and gradient stability. 
Pseudo-labeling in collaborative training applies confidence thresholds specific to each branch: $\theta_1=0.5$ for ViT-generated labels and a more stringent $\theta_2=0.8$ for CNN. 
These empirically chosen thresholds ensure that only high-confidence pseudo-labels contribute to model training.

\section{Experiments}
\label{S5}
 
\subsection{Experimental Setup}

\textbf{Classification Task}.    
We carried out two classification tasks: (1) binary classification between CN-D and CN-N on the target domain (NBOLD), and (2) three-category classification among CI, CN-D, and CN-N on the target domain (NBOLD). 
For binary classification, the mean and standard deviation of the five evaluation metrics were computed across the five cross-validation folds. 
For three-category classification, we employed the widely used one-vs-rest strategy~\cite{aly2005one}. 
Predictions across all cross-validation folds were aggregated to calculate the overall accuracy (ACC) and sensitivity (SEN), along with per-class ACC and SEN scores.

\textbf{Evaluation Metric}.   
To comprehensively assess model performance, we employed several widely used evaluation metrics, including area under the receiver operating characteristic curve (AUC), accuracy (ACC), sensitivity (SEN), specificity (SPE), and F1-score (F1). 
All experiments were conducted under a five-fold cross-validation (CV) strategy for the target data.   
Specifically, target subjects within each class were randomly partitioned into five (approximately equal-sized) subsets. 
In each iteration, one subset was held out as the test set, while the remaining four subsets were used for training. 
This CV procedure was repeated five times with different random seeds to further minimize variance,  and to reduce potential bias introduced by sample partitioning.

\textbf{Competing Methods}. 

We compare our CDA with eleven competition methods, including three traditional domain adaptation methods (using 1,373-dimensional handcrafted features
extracted via FreeSurfer), as well as eight state-of-the-art deep domain adaptation methods with 3D MRI scans as input. 
We mainly use the default setting of these competing methods and make a concerted effort to ensure that the network architecture and hyperparameters are comparable to the proposed CDA.

(1) Transfer Component Analysis (\textbf{TCA})~\cite{pan2010domain} is a domain adaptation technique that seeks to learn a shared subspace between source and target domains by minimizing the Maximum Mean Discrepancy (MMD). This approach utilizes kernel methods to map data into a Reproducing Kernel Hilbert Space (RKHS), where the MMD criterion is applied to reduce distributional differences between domains. 
Its goal is to identify transfer components that align the marginal distributions of both domains, enabling the application of classifiers (trained on the source domain) to the target domain. 

(2) Joint Distribution Adaptation (\textbf{JDA})~\cite{long2013transfer} 
aims to align both marginal and conditional distributions between source and target domains. 
It integrates MMD into feature learning to find  robust shared representations. 
It begins with dimensionality reduction via Principal Component Analysis (PCA), then iteratively minimizes the distribution discrepancy by aligning the marginal distribution and the class-conditional distributions using MMD. 
For unlabeled target data, JDA employs pseudo-labels generated from a classifier trained on source data, and iteratively refines these pseudo-labels along with the feature transformation matrix, to improve alignment of conditional distributions and overall adaptation performance.

(3) Scatter Component Analysis (\textbf{SCA})~\cite{ghifary2016scatter} is a kernel-based feature learning method designed for both domain adaptation and domain generalization. It learns a shared latent space that aligns samples from different domains while preserving strong class separability. 
SCA achieves this by simultaneously maximizing total scatter and between-class scatter—enhancing overall variance and inter-class distinction—and by minimizing within-class scatter and domain scatter—reducing intra-class variability and domain discrepancy. 
Operating in a RKHS via the kernel trick, SCA solves a generalized eigenvalue problem to efficiently learn domain-invariant representations that generalize well across domains.

(4) Deep Adaptation Network (\textbf{DAN})~\cite{long2015learning} is a deep transfer learning framework that aligns feature distributions across domains by applying multi-kernel Maximum Mean Discrepancy (MK-MMD) to multiple layers of a deep neural network. Built on standard CNN architectures, DAN reduces both marginal and conditional distribution shifts between the source and target domains through multi-layer feature adaptation. By embedding MK-MMD into specific hidden layers, DAN enables the learning of transferable representations that generalize well to unlabeled target data.

(5) Domain-Adversarial Neural Network (\textbf{DANN})~\cite{ganin2015unsupervised} aims to learn domain-invariant and task-discriminative features via adversarial training. It consists of three main components: a feature extractor, a label predictor for source supervision, and a domain classifier that distinguishes between source and target domains. A key innovation is the gradient reversal layer, which inverts gradients from the domain classifier during backpropagation, encouraging the feature extractor to learn representations that are indistinguishable across domains. This adversarial objective promotes effective domain alignment while preserving label prediction performance.

(6) \textbf{DeepCoral}~\cite{sun2016deep} 
is a UDA method that aligns second-order statistics of source and target feature distributions to reduce domain shift. It introduces the CORAL (Correlation Alignment) loss, which minimizes the Frobenius norm of the difference between the covariance matrices of source and target features. By incorporating this loss alongside the standard classification loss, DeepCoral encourages the network to learn domain-invariant representations without requiring target labels.

(7) Adversarial Discriminative Domain Adaptation (\textbf{ADDA})~\cite{tzeng2017adversarial} is a UDA method that spatially aligns target domain features with source domain features through adversarial learning. 
Its operation is divided into three stages: 1) pre-training the source encoder and classifier using labeled source data; 2) freezing the source model and adversarially training a target encoder with a domain discriminator to align source and target feature distributions;
3) directly applying the source classifier to adaptive target features. 

(8) Dynamic Adversarial Adaptation Network (\textbf{DAAN})~\cite{yu2019transfer} aims to dynamically balance global and class-conditional (local) feature alignment through adversarial learning. It comprises three main modules: a label classifier for source supervision, a global domain discriminator for aligning marginal distributions, and multiple local subdomain discriminators for aligning class-conditional distributions. Its dynamic adversarial factor adaptively weighs the contribution of global versus local alignment during training. This factor is computed based on the losses of the global and local discriminators, allowing the model to adjust the alignment strategy in response to training dynamics and domain shift characteristics.

(9) Batch Nuclear-norm Maximization (\textbf{BNM})~\cite{cui2020towards}
is a UDA method, designed to improve prediction discriminability and diversity in the absence of target labels. Instead of relying on traditional entropy minimization, BNM enhances feature representation by maximizing the nuclear norm of the batch-wise prediction probability matrix. This encourages predictions to span multiple classes (diversity) while remaining confident (discriminability), leading to more transferable decision boundaries. BNM is typically integrated into deep models (e.g., ResNet18, ResNet50) and operates without modifying the feature extractor architecture. 

(10) Deep Subdomain Adaptation Network (\textbf{DSAN})~\cite{zhu2020deep} 
performs fine-grained alignment by minimizing distribution discrepancies at the subdomain (class-specific) level rather than only aligning global distributions. DSAN introduces Local Maximum Mean Discrepancy (LMMD) to measure and reduce the distance between subdomains of the same class across source and target domains. This is achieved by assigning soft pseudo-labels to unlabeled target samples and computing class-conditional alignment losses. By jointly optimizing classification loss and LMMD loss over multiple feature layers, DSAN enables discriminative and domain-invariant representation learning. This strategy improves adaptation performance, particularly in tasks where intra-class variability and inter-class similarity pose significant challenges.

(11) Source Hypothesis Transfer++ (\textbf{SHOT++})~\cite{liang2021source} is designed for scenarios where only the trained source model (not  source data) is available during adaptation. 
It operates by freezing the source classifier and adapting only the target feature extractor to align target features with the fixed source hypothesis. It consists of three  components: information maximization, which encourages confident and diverse target predictions; self-supervised pseudo-label refinement, which improves feature clustering by enforcing consistency; and label propagation that divides target samples into high- and low-confidence groups and propagates label information from confident to uncertain samples.

\begin{table}[!t]
\setlength{\belowdisplayskip}{0pt}
\setlength{\abovedisplayskip}{0pt}
\setlength{\abovecaptionskip}{0pt}
\setlength{\belowcaptionskip}{0pt}
\renewcommand{\arraystretch}{1.1}
\setlength\tabcolsep{1pt}
\small
\centering
\caption{Results achieved by different domain adaptation methods (mean±standard deviation) for CN-D vs. CN-N classification.}
\resizebox{\linewidth}{!}{
\begin{tabular}{l|cc cc c|c}
\toprule
Method & AUC~(\%) & ACC~(\%) & SEN~(\%) & SPE~(\%) & F1~(\%) & $p$-value \\ 
\midrule
TCA & 49.62$\pm$1.95 & 50.75$\pm$1.70 & 63.88$\pm$6.13 & 35.29$\pm$8.32 & 39.18$\pm$6.53 & 0.0001\\ 
JDA & 54.15$\pm$1.38 & 54.35$\pm$1.12 & 62.03$\pm$1.45 & 46.27$\pm$4.01 & 50.54$\pm$4.61 & 0.0002\\ 
SCA & 51.95$\pm$2.76 & 52.10$\pm$1.66 & 58.26$\pm$8.26 & 37.84$\pm$17.14 & 32.11$\pm$4.45 & 0.0005\\
\hline
DAN & 68.87$\pm$0.38 & 62.00$\pm$3.08 & 67.15$\pm$3.14 & 50.00$\pm$13.69 & 71.23$\pm$1.69 & 0.0012\\
DAAN & 67.37$\pm$10.98 & 65.28$\pm$3.38 & 68.89$\pm$7.54 & 42.73$\pm$21.97 & 55.15$\pm$16.13 & 0.0327\\
DeepCoral & 68.87$\pm$0.65 & 62.00$\pm$1.00 & 67.86$\pm$3.09 & 48.33$\pm$8.07 & 71.40$\pm$1.07 & 0.0079\\
ADDA & 52.39$\pm$3.64 & 51.53$\pm$13.94 & 64.45$\pm$10.08 & 36.07$\pm$12.35 & 53.78$\pm$9.80 & 0.0002 \\
DANN & 57.88$\pm$9.46 & 57.93$\pm$4.58 & 66.57$\pm$7.43 & 46.16$\pm$10.15 & 47.49$\pm$8.60  & 0.0003\\
BNM & 68.87$\pm$0.78 & 60.00$\pm$3.06 & 67.15$\pm$15.42 & 43.33$\pm$17.10 & 71.14$\pm$1.59 & 0.0063\\
DSAN & 68.81$\pm$0.25 & 61.00$\pm$2.92 & 68.57$\pm$3.61 & 43.33$\pm$3.73 & 71.07$\pm$2.40 & 0.0079\\
SHOT++ & 54.00$\pm$10.13 & 50.37$\pm$8.21 & 54.76$\pm$12.27 & 56.67$\pm$10.16 & 52.37$\pm$10.57 & 0.0056\\
\hline
CDA (Ours) & \textbf{71.51$\pm$6.82} & \textbf{70.73$\pm$8.59} & \textbf{69.29$\pm$8.32} & \textbf{73.09$\pm$12.42} & \textbf{74.99$\pm$10.62} & -- \\
\bottomrule
\end{tabular}
}
\label{binary_classification}
\end{table}

\subsection{Results of Binary Classification}
Table~\ref{binary_classification} reports the comparative performance of multiple domain adaptation approaches on the binary classification task distinguishing cognitively normal individuals with and without depression (CN-D vs. CN-N). 
We also report the $p$-values from paired sample $t$-test to assess whether the performance differences between the proposed CDA and specific competing methods are statistically significant (threshold: $p < 0.05$). 
From Table~\ref{binary_classification}, we have the following observations.

\emph{First}, traditional unsupervised domain adaptation (UDA) methods based on handcrafted MRI features, such as TCA, SCA, and JDA,  show substantially inferior performance (AUC: 49.62\%–54.15\%, ACC: 50.75\%–54.35\%) when compared to the end-to-end deep learning models. 
omical variations in 3D structural MRI. 
This performance gap likely stems from the limited capacity of handcrafted features to capture the complex neuroanatomical variations present in 3D structural MRI. 
In contrast, deep models operating directly on raw MRI volumes achieve markedly better results, highlighting the superiority of learned hierarchical representations for LLD detection. 
\emph{Second}, among MMD-based deep learning approaches, DAN and DSAN demonstrate effective alignment via multi-layer and subdomain-level feature adaptation. 
However, their performance remains consistently inferior to that of CDA, highlighting the added benefits of collaborative and bidirectional adaptation strategies.  
\emph{Third}, adversarial learning-based methods such as DAAN also performed well in terms of AUC and ACC, indicating a relatively strong capacity to handle domain shifts. 
However, other adversarial approaches like ADDA lagged significantly behind across nearly all metrics, especially in terms of SPE and F1, suggesting unstable performance and limited generalization to the target domain. 
\emph{Fourth}, information-theoretic models like BNM and SHOT++ aim to improve class separability. While BNM achieves competitive performance among deep UDA methods, SHOT++ lags behind, highlighting substantial differences in their generalization capabilities. 
\emph{Finally}, the CDA framework achieves the highest performance across all metrics, outperforming all competing methods in this task. 
Statistical significance testing further confirms that CDA's improvements are significant ($p < 0.05$) compared to most competing methods, suggesting the robustness and reliability of CDA in addressing domain shift challenges in cross-site MRI analysis.

\begin{table}[!t]
\setlength{\belowdisplayskip}{0pt}
\setlength{\abovedisplayskip}{0pt}
\setlength{\abovecaptionskip}{0pt}
\setlength{\belowcaptionskip}{0pt}
\renewcommand{\arraystretch}{1.1}
\setlength\tabcolsep{1pt}
\small
\centering
\caption{Accuracy (ACC) results achieved by different methods (mean±standard deviation) in the task of three-category classification (i.e., CI vs. CN-D vs. CN-N) on the target data.}
\resizebox{\linewidth}{!}{
\begin{tabular}{l|cccc|c}
\toprule
Method & ACC~(\%) & ACC$_{\text{CN-N}}$~(\%) & ACC$_{\text{CN-D}}$~(\%) & ACC$_{\text{CI}}$~(\%) & $p$-value \\ 
\midrule
TCA & 40.81$\pm$7.21 & 57.98$\pm$6.91 & 45.82$\pm$8.65 & 67.83$\pm$7.28 & 0.0007\\ 
JDA & 44.22$\pm$5.57 & 60.97$\pm$3.09 & 52.71$\pm$8.20 & 74.76$\pm$7.21 & 0.0006\\ 
SCA & 45.09$\pm$13.9 & \textbf{69.66$\pm$5.43} & 59.87$\pm$9.67 & 72.92$\pm$8.07 & 0.0180\\
\hline
DAN & 34.59$\pm$17.44 & 60.73$\pm$16.56 & 49.33$\pm$9.41 & 59.12$\pm$12.65 & 0.0023\\
DANN & 48.83$\pm$11.71 & 62.19$\pm$9.34 & 54.84$\pm$7.59 &80.64$\pm$9.55 & 0.0020\\
DeepCoral & 36.17$\pm$6.96 & 59.72$\pm$7.10 & 43.69$\pm$2.18 & 68.93$\pm$11.42 & 0.0004\\
ADDA & 44.13$\pm$16.89 & 63.77$\pm$15.98 & 52.33$\pm$6.47 & 72.17$\pm$17.78 & 0.0362 \\
DAAN & 50.09$\pm$5.86 & 68.57$\pm$4.88 & 56.54$\pm$4.44 & 80.70$\pm$2.75 & 0.0449 \\
BNM & 34.00$\pm$4.52 & 67.33$\pm$4.42 & 48.00$\pm$5.42 & 48.67$\pm$9.57 & 0.0000\\ 
DSAN & 43.33$\pm$7.30 & 61.33$\pm$16.14 & 52.67$\pm$2.49 & 72.67$\pm$8.54 & 0.0007\\
SHOT++ & 45.57$\pm$7.87 & 53.87$\pm$13.38 & 49.43$\pm$16.63 & \textbf{82.58$\pm$3.57} & 0.0039 \\
\hline
CDA (Ours) & \textbf{50.79 $\pm$8.93} & 59.49$\pm$6.75 & \textbf{64.22$\pm$5.32} & 64.50$\pm$10.00 & -- \\
\bottomrule
\end{tabular}
}
\label{ACC_threecategory}
\end{table}

\subsection{Results of Three-Category Classification}
For CI vs. CN-D vs. CN-N classification, we report the ACC and SEN results 
achieved by different methods for each category in Tables~\ref{ACC_threecategory} and~\ref{SEN_threecategorys}, respectively.

As shown in Table~\ref{ACC_threecategory}, our CDA method achieves the highest overall ACC result as well as the highest ACC value for the CN-D category, outperforming all competing models. 
CDA also shows more balanced performance across the three categories, with ACC values of 59.49\% for CN-N, 64.22\% for CN-D, and 64.50\% for CI.  
The three deep learning methods (DANN, DAAN, and SHOT++) yield very imbalanced results among the three categories, with lower overall accuracy compared with CDA. 
The results in Table~\ref{SEN_threecategorys}  suggest that our CDA obtains the highest overall SEN and highest SEN for CN-N 
and comparable performance for CN-D, but the sensitivity for CI is not good. Notably, several methods such as DANN, DAAN, and DSAN obtain relatively high SEN scores for CN-D, but their performance on CI and CN-N is substantially lower, which again reflects class imbalance in predictions.  
Although CDA achieves reasonable SEN for CN-D and CN-N, the SEN value for the CI class remains suboptimal. 
This may be due to the limited number of CI subjects (only 17), and the even smaller representation of this class in each cross-validation test fold. 
Such data imbalance likely hinders the model's ability to generalize effectively in the multi-category classification setting.

Comparing the three-category classification results in Tables~\ref{ACC_threecategory}–\ref{SEN_threecategorys} with the binary classification results in Table~\ref{binary_classification}, we observe a notable decline in overall performance when transitioning from binary to multi-class classification. 
This performance drop is largely attributable to the increased difficulty in accurately identifying the newly introduced CI category.

\begin{table}[!t]
\setlength{\belowdisplayskip}{0pt}
\setlength{\abovedisplayskip}{0pt}
\setlength{\abovecaptionskip}{0pt}
\setlength{\belowcaptionskip}{0pt}
\renewcommand{\arraystretch}{1.1}
\setlength\tabcolsep{1pt}
\small
\centering
\caption{Sensitivity (SEN) results achieved by different methods (mean±standard deviation) in the task of three-category classification (i.e., CI vs. CN-D vs. CN-N) on the target data.} 
\resizebox{\linewidth}{!}{
\begin{tabular}{l|cccc|c}
\toprule
Method & SEN~(\%) & SEN$_{\text{CN-N}}$~(\%) & SEN$_{\text{CN-D}}$~(\%) & SEN$_{\text{CI}}$~(\%) & $p$-value \\ 
\midrule
TCA & 34.22$\pm$2.720 & 21.43$\pm$11.19 & 59.58$\pm$10.73 & 21.67$\pm$9.44 & 0.0430\\ 
JDA & 37.64$\pm$7.88 & 39.64$\pm$19.90 & 53.28$\pm$1.88 & 20.00$\pm$1.87 & 0.0887\\ 
SCA & 33.81$\pm$0.95 & 32.50$\pm$20.22 & 63.29$\pm$24.95 & \textbf{43.33$\pm$16.16} & 0.0194\\
\hline
DAN & 35.24$\pm$2.33 & 28.57$\pm$17.25 & 37.14$\pm$15.71 & 40.00$\pm$8.99 & 0.0375\\
DANN & 37.95$\pm$11.14 & 22.86$\pm$11.03 & 69.32$\pm$14.33 & 21.67$\pm$9.63 & 0.0487\\
DeepCoral & 31.28$\pm$7.69 & 21.07$\pm$11.32 & 46.10$\pm$21.97 & 26.67$\pm$7.59 & 0.0418\\
ADDA & 33.33$\pm$0.34 & 20.00$\pm$4.89 & 60.00$\pm$4.53 & 20.00$\pm$2.67 & 0.0041 \\
DAAN & 35.41$\pm$3.77 & 25.36$\pm$6.59 & \textbf{70.88$\pm$14.93} & 10.00$\pm$2.00 & 0.0588\\
BNM & 29.58$\pm$4.90 & 7.50$\pm$10.00 & 41.25$\pm$14.03 & 40.00$\pm$22.61 & 0.0078\\ 
DSAN & 31.11$\pm$1.88 & 20.00$\pm$21.79 & 70.00$\pm$23.18 & 3.33$\pm$6.67 & 0.0084\\
SHOT++ & 34.20$\pm$5.31 & 42.86$\pm$15.75 & 59.74$\pm$12.14 & 6.67$\pm$13.33 & 0.0189 \\
\hline
CDA (Ours) & \textbf{42.22$\pm$6.52} & \textbf{50.67$\pm$3.79} & 63.33 $\pm$4.52 & 12.67$\pm$1.33 & -- \\
\bottomrule
\end{tabular}
}
\label{SEN_threecategorys}
\end{table}

\section{Discussion}
\label{S6}
\subsection{Ablation Study}
To evaluate the contribution of each stage  in our CDA framework, we compare CDA with four ablated variants: \textbf{CDA\_S$_1$}, \textbf{CDA\_S$_{12}$}, \textbf{CDA\_S$_{13}$}, and \textbf{CDA\_S$_{23}$}.
Specifically, {CDA\_S$_1$} involves only Stage 1 (supervised training on labeled source data). 
{CDA\_S$_{12}$} incorporates both Stage 1 and Stage 2 (self-supervised target feature adaptation). 
{CDA\_S$_{13}$} combines Stage 1 with Stage 3 (collaborative training on unlabeled target data). 
{CDA\_S$_{23}$} utilizes only Stage 2 and Stage 3, without supervised source training. 
All variants use the same backbone and hyperparameters to ensure a fair comparison.

\begin{figure}[!t]
\setlength{\belowdisplayskip}{0pt}
\setlength{\abovedisplayskip}{0pt}
\setlength{\abovecaptionskip}{0pt}
\setlength{\belowcaptionskip}{0pt}
\centering
\includegraphics[width=0.47\textwidth]{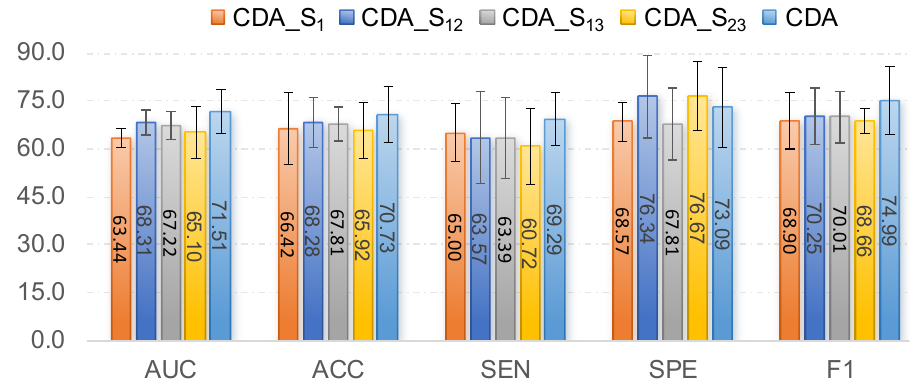}
\caption{{Ablation study comparing CDA and its four variants in the task of CN-D vs. CN-N classification.}}
\label{Ablation_Binary}
\end{figure}

As shown in Fig.~\ref{Ablation_Binary}, the CDA achieves the overall best performance. 
For instance, CDA yields an AUC of 71.51\%, which is higher than all baselines. Similarly, CDA obtains the highest F1 of 74.99\%, outperforming its variants by 1.90\%–5.70\%, demonstrating its advantage in handling class imbalance and improving discriminative capacity. 
Notably, both CDA\_S$_{12}$ and CDA\_S$_{13}$ outperform CDA\_S$_1$ in most cases, 
demonstrating that incorporating either self-supervised adaptation (Stage 2) or collaborative training (Stage 3) leads to improved generalization over supervised learning alone. 
The full CDA framework, which integrates all three stages, achieves the most substantial performance gains, validating the complementary benefits of these strategies. Moreover, CDA\_S$_{23}$ also shows competitive results, even without supervised source training, further emphasizing the effectiveness of target-domain self-supervision and cross-branch collaboration. 
These findings confirm the efficacy of CDA's staged components for LLD identification.

\subsection{Effect of Dual-Branch Backbone Architecture}

\begin{figure}[!t]
\centering
\includegraphics[width=0.47\textwidth]{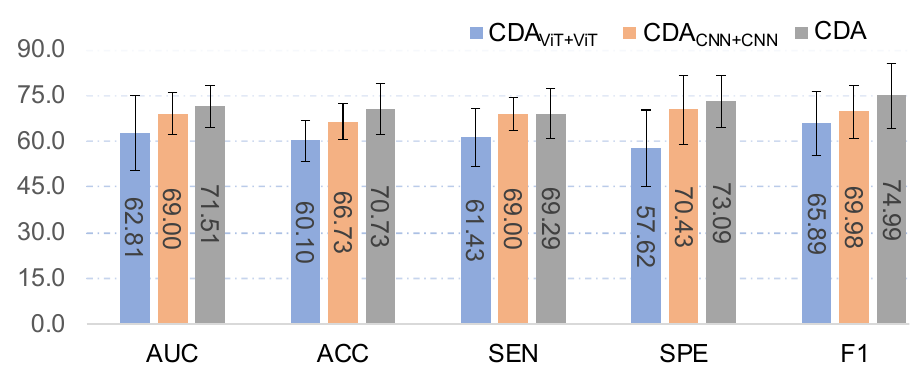}
\caption{Performance comparison of CDA and its two variants (i.e., $\text{CDA}_{\text{ViT+ViT}}$ and $\text{CDA}_{\text{CNN+CNN}}$) employing different backbone architectures for CN-D vs. CN-N classification.}
\label{dif_Architecture}
\end{figure}

To investigate the effect of different backbone architectures on performance, we conduct a comparative study across three configurations: \textbf{$\text{CDA}_{\text{CNN+CNN}}$} with two CNN branches, \textbf{$\text{CDA}_{\text{ViT+ViT}}$} with two ViT branches, and our proposed hybrid ViT+CNN used in CDA. 
The experimental results are presented in Fig.~\ref{dif_Architecture}. 
As shown in this figure, the hybrid CDA model consistently outperforms 
 both homogeneous variants across all evaluation metrics. 
Specifically, it achieves the highest AUC, ACC, SEN, SPE, and F1, significantly surpassing both  $\text{CDA}_{\text{CNN+CNN}}$,$\text{CDA}_{\text{ViT+ViT}}$.  
These results suggest the benefit of integrating global and local representations within a unified framework. 
The ViT branch contributes strong global context modeling, while the CNN branch captures fine-grained local texture information. 
This combination enhances feature diversity and leads to more effective cross-domain alignment. 
Overall, the empirical evidence validates the effectiveness of the hybrid ViT–CNN architecture in CDA, demonstrating its superiority over purely homogeneous designs.

\subsection{Hyperparameter Analysis}
\begin{figure}[!t]
\setlength{\belowdisplayskip}{0pt}
\setlength{\abovedisplayskip}{0pt}
\setlength{\abovecaptionskip}{0pt}
\setlength{\belowcaptionskip}{0pt}
\centering\includegraphics[width=0.47\textwidth]{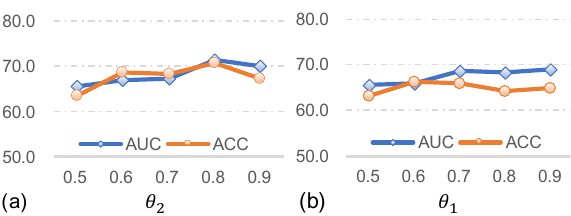}
\caption{Results of CDA with different threshold combinations in the task of CN-D vs. CN-N classification.} 
\label{Threshold_vitcnn}
\end{figure}

For collaborative training on the unlabeled target data, the selection of threshold values in Eqs.~\eqref{eq_lossVITtoCNN}-\eqref{eq_lossCNNtoVIT} plays a crucial role in determining the overall training effectiveness.
Fig.~\ref{Threshold_vitcnn} shows the impact of varying thresholds $\theta_{\text{2}}$ and $\theta_{\text{1}}$ on model performance.  
In Fig.~\ref{Threshold_vitcnn}~(a), we fix $\theta_{\text{1}}=0.5$ and vary $\theta_{\text{2}}$ from 0.5 to 0.9. 
As $\theta_{\text{2}}$ increases, both AUC and ACC improve steadily, peaking at $\theta_{\text{2}} = 0.8$, and slightly declining at 0.9, suggesting that the model performs  robustly near $\theta_{\text{2}} = 0.8$.  
In Fig.~\ref{Threshold_vitcnn}~(b), we fix $\theta_{\text{2}}=0.5$ and adjust $\theta_{\text{1}}$ over the same range. 
With fixed $\theta_2$, CDA with a small $\theta_1$ (e.g., $0.5$) yields worse performance.  
The likely reason is that a small $\theta_1$ leads the model to accept low-confidence pseudo-labels generated by ViT, which hinders effective knowledge transfer from ViT to CNN during training.

\subsection{Influence of Collaborative Training Strategy}
\begin{figure}[!t]
\setlength{\belowdisplayskip}{0pt}
\setlength{\abovedisplayskip}{0pt}
\setlength{\abovecaptionskip}{0pt}
\setlength{\belowcaptionskip}{0pt}
\centering
\includegraphics[width=0.47\textwidth]{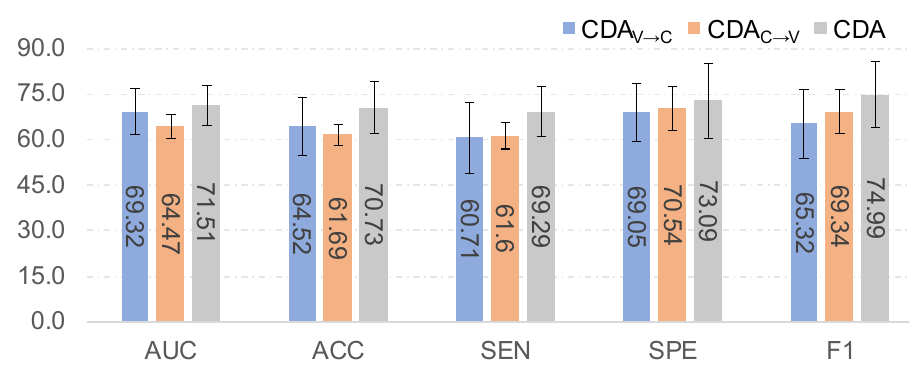}
\caption{{Comparison of CDA and its two variants with different collaborative training strategies for CN-D vs. CN-N classification.}}
\label{Cotrain}
\end{figure}

To validate the benefits of collaborative learning in CDA, we conduct an ablation study comparing CDA with two variants that use different collaborative training strategies: \textbf{CDA$_{V \rightarrow C}$}, and \textbf{CDA$_{C \rightarrow V}$}. 
Both variants share the same first two stages as the original CDA framework. 
In the third stage of CDA$_{V \rightarrow C}$, the ViT branch generates pseudo labels from its predictions on unlabeled target data to guide the CNN branch by minimizing the loss $\mathcal{L}_{{V}\rightarrow{C}}$.  Conversely, in CDA$_{C \rightarrow V}$, the CNN branch generates pseudo labels to supervise the ViT branch by minimizing $\mathcal{L}_{{C}\rightarrow{V}}$. 
Results in Fig.~\ref{Cotrain} show that the full CDA framework achieves the highest results across all metrics. 
For instance, CDA$_{V \rightarrow C}$ achieves 69.32\% AUC and 64.52\% ACC, indicating limited adaptation when only the ViT branch provides supervision. 
CDA$_{C \rightarrow V}$ performs even worse, showing further declines in both metrics. 
Notably, CDA$_{V \rightarrow C}$ still underperforms the full CDA by 2.19\% in AUC, indicating the complementary contributions of both modality-specific branches and the advantage of their integration. 
These observations demonstrate that one-directional supervision is suboptimal due to the asymmetric nature and learning capacity of CNN and ViT architectures. In contrast, the fully cooperative training strategy—wherein the two branches mutually exchange pseudo-labels and iteratively refine their predictions—demonstrated superior performance across all evaluation metrics. This significant gain validates that the bidirectional pseudo-labeling encourages mutual enhancement and alignment between the global (ViT) and local (CNN) feature representations. 
The collaborative learning strategy plays a critical role in enhancing domain adaptation by balancing representation diversity and decision consistency.

\subsection{Effect of Inference Model
}
\label{S6_EffectInference}

\begin{figure}[!t]
\setlength{\belowdisplayskip}{0pt}
\setlength{\abovedisplayskip}{0pt}
\setlength{\abovecaptionskip}{0pt}
\setlength{\belowcaptionskip}{0pt}
\centering
\includegraphics[width=0.47\textwidth]{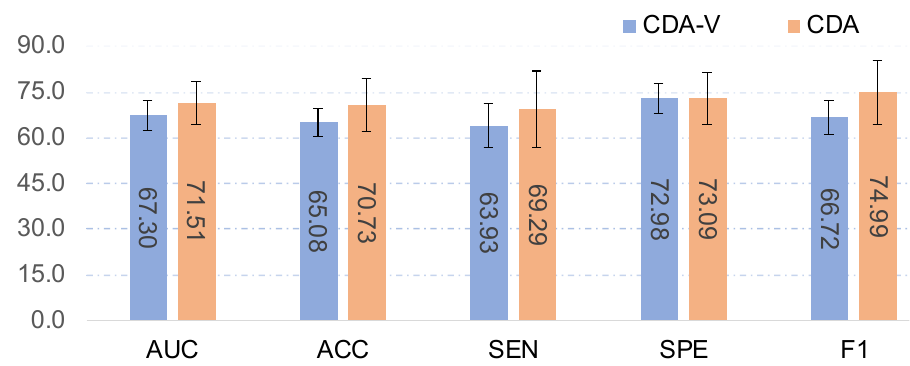}
\caption{{Comparison between CDA (with CNN for inference) and CDA-V (with ViT for inference) in the task of CN-D vs. CN-N classification.}}
\label{IntferViTCNN}
\end{figure}

While the main experiments utilize the trained CNN model for inference in CDA, we also evaluate a variant employing the ViT model for inference, referred to as \textbf{CDA-V}. 
The results, reported in Fig.~\ref{IntferViTCNN}, show that CDA-V achieves 67.30\% AUC, 65.08\% ACC, 63.93\% SEN, 72.98\% SPE, and 66.72\% F1, all lower than CDA's corresponding metrics.  
The experimental results suggest that CNN-based local representations may provide superior discriminative capacity compared to ViT's global modeling when applied to target domain inference. Although ViT is known for capturing global context, its standalone performance appears limited in this setting, potentially due to a lower inductive bias and greater sensitivity to limited data. 
In contrast, CNN strengthened through collaborative training and pseudo-label refinement effectively captures local features while benefiting from the stability introduced by joint supervision during collaborative training. 
These findings strongly support the core hypothesis of CDA: the collaborative training framework enables CNN to harness ViT's generalization strengths while preserving robust and precise decision boundaries through local receptive fields, thus improving cross-domain classification.

\subsection{Influence of Encoder Fine-Tuning}
\begin{figure}[!t]
\setlength{\belowdisplayskip}{0pt}
\setlength{\abovedisplayskip}{0pt}
\setlength{\abovecaptionskip}{0pt}
\setlength{\belowcaptionskip}{0pt}
\centering
\includegraphics[width=0.47\textwidth]{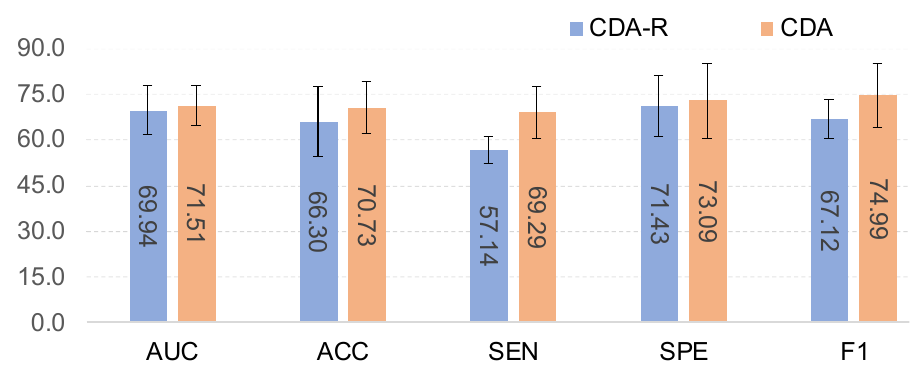}
\caption{{Comparison between CDA and its variant CDA-R with different encoder fine-tuning strategy for CN-D vs. CN-N classification.}} 
\label{CNNfine-Tune}
\end{figure}

To assess the necessity of asymmetric design in Stage 2 for target-source feature alignment (see Fig.~\ref{CDA}~(b)), we carry out a controlled experiment by interchanging the roles of ViT and CNN branches during discrepancy optimization. 
In the original CDA setting, we freeze the ViT encoder ($E_V$) while updating the two classifiers to maximize their discrepancy. 
We then fine-tune the CNN encoder ($E_C$) while keeping the classifiers fixed, minimizing their discrepancy to achieve stable feature alignment. 
To evaluate the sensitivity of this design, we reverse the roles of the two branches: we first freeze the CNN encoder while updating the classifiers, and then fine-tune the ViT encoder with the classifiers frozen. We refer to this variant as \textbf{CDA-R}. 
The resulting performance of CDA and CDA-R is  reported in  Fig.~\ref{CNNfine-Tune}. 
As shown, CDA-R exhibits a notable decline compared to CDA, particularly in SEN, which drops to 57.14\%. 
This implies that CNN lacks the global capacity needed for effective discrepancy maximization, while ViT is more sensitive to noisy pseudo-labels during alignment. 
These results emphasize the importance of CDA's  asymmetric design: ViT encoder promotes boundary diversification, while CNN encoder ensures stable alignment, together enabling effective cross-domain adaptation.

\subsection{Influence of Source Domain} 
We use NCODE as the source domain in the main experiments. 
To further assess the adaptability of our CDA with different source domains, we perform a comparative study using ADNI~\cite{jack2008alzheimer} as the source domain, while maintaining NBOLD as the target domain. 
We refer to this method as \textbf{CDA-ADNI}. 
For CDA-ADNI, the baseline T1-weighted MRI scans from CN and AD groups in ADNI-1 and ADNI-2 are used as source data, including 359 AD and 436 CN subjects.

\begin{figure}[!t]
\setlength{\belowdisplayskip}{0pt}
\setlength{\abovedisplayskip}{0pt}
\setlength{\abovecaptionskip}{0pt}
\setlength{\belowcaptionskip}{0pt}
\centering
\includegraphics[width=0.47\textwidth]{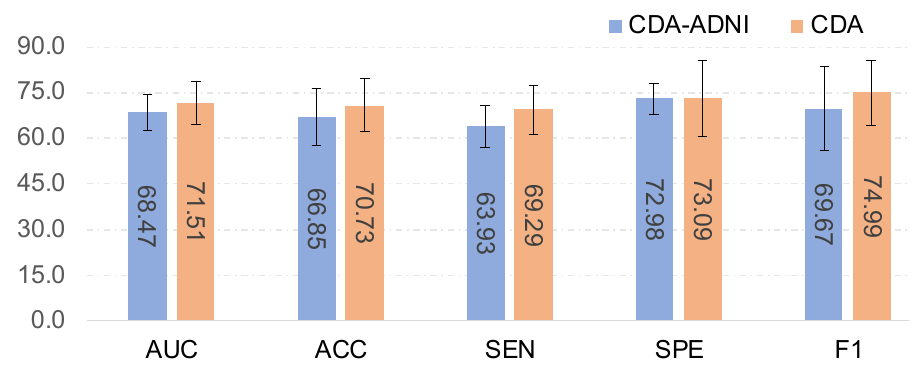}
\caption{Results of our CDA (with NCODE as the source domain) and its variant CDA-ADNI (with ADNI as the source domain) for  CN-D vs. CN-N classification on the target NBOLD domain.}
\label{Crossdata}
\end{figure}

As summarized in Fig.~\ref{Crossdata}, training CDA with NCODE as the source domain consistently outperforms its counterpart trained on ADNI (CDA-ADNI) across all evaluation metrics. Specifically, CDA achieves improvements of +3.04\% in AUC, +3.88\% in ACC, +5.36\% in SEN, +0.11\% in SPE, and +5.32\% in F1 relative to CDA-ADNI. 
This performance gap is largely attributable to the greater domain similarity between NCODE and the target NBOLD dataset, including more consistent imaging protocols, better demographic alignment, and a shared diagnostic taxonomy.  
Despite this domain gap, CDA-ADNI still delivers competitive results, owing to the substantial volume of training data it provides. 
Notably, the ADNI dataset includes 795 MRI scans, offering rich structural variability and a broader representation space that serve as strong auxiliary supervision for downstream tasks. 
These findings highlight a key insight: while domain proximity facilitates feature transferability and decision boundary alignment, large-scale auxiliary datasets enhance representation robustness and mitigate overfitting to domain-specific artifacts. 
The proposed CDA framework adapts well to different source data conditions. 
It takes advantage of semantic similarity when available (as with NCODE) and benefits from large, diverse datasets when similarity is lower (as with ADNI), showing strong flexibility in cross-domain neuroimaging classification.

\subsection{Impact of Encoder Pretraining}
\begin{figure}[!t]
\setlength{\belowdisplayskip}{0pt}
\setlength{\abovedisplayskip}{0pt}
\setlength{\abovecaptionskip}{0pt}
\setlength{\belowcaptionskip}{0pt}
\centering
\includegraphics[width=0.47\textwidth]{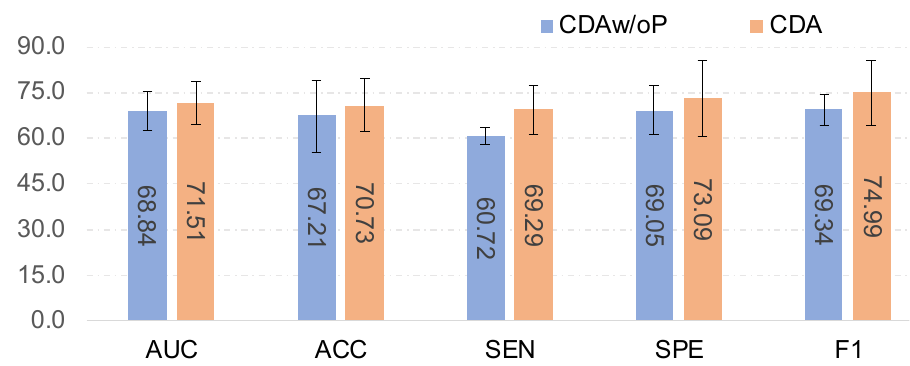}
\caption{{Comparison between CDA and its variant (CDAw/oP) without encoder pretraining for CN-D vs. CN-N classification.}}
\label{Pretrain}
\end{figure}

In the main experiments, we employ pretrained encoders ($E_V$ and $E_C$) in ViT and CNN,  respectively. 
Specifically, $E_V$ is pretrained on a combination of three datasets (IXI, OASIS-3, and BRATS)~\cite{kunanbayev2024training}, while $E_C$ is pretrained on ADNI~\cite{zhang2025brain}. 
To assess the impact of this encoder pretraining strategy, we compare our CDA with a baseline variant, \textbf{CDAw/oP}, which uses randomly initialized encoders without any pretraining. 
The results of the two methods are reported in Fig.~\ref{Pretrain}. 
As illustrated in this figure, CDA with encoder pretraining consistently outperforms CDAw/oP on downstream tasks: AUC increases by 2.67\%, ACC by 3.52\%, SEN by 8.57\%, SPE by 4.04\%, and F1 score by 5.65\%.  
These performance gains validate the effectiveness of our pretraining strategy, which leverages structural prior knowledge to learn more transferable and domain-invariant representations.  
The findings are consistent with prior work~\cite{zhang2025brain}, while further extending its applicability to domain adaptation scenarios. 
By encoding fundamental anatomical patterns in MRI during pretraining, the model gains robustness against inter-domain variability and becomes less dependent on large labeled samples in the source domain.

\subsection{Limitations and Future Work} 
Although our CDA demonstrates strong performance, several limitations remain to be addressed. 
\emph{First}, the current model relies on a fixed backbone architecture (ViT and CNN) without further architectural optimization for the LLD detection task. 
Future work could explore the integration of more advanced backbones, such as hybrid transformer-convolutional architectures or domain-specific neural operators, to better capture complex neuroanatomical patterns in LLD cohorts. 
\emph{Second}, while CDA achieves robust cross-domain performance, its generalization to entirely unseen imaging sites with more extreme distribution shifts remains to be evaluated.
In future studies, incorporating domain generalization strategies or meta-learning paradigms may further enhance adaptability. 
\emph{Lastly}, multimodal neuroimaging data (e.g., functional MRI, DTI) could provide additional complementary information, and integrating such data into CDA may further improve diagnostic accuracy, which will also be our future work.

\section{Conclusion}
\label{S7}
This study introduces a CDA framework for automated LLD identification using T1-weighted MRI scans. 
CDA combines a ViT and a CNN to extract global and local brain features. 
Its training consists of three stages: supervised learning on source data, self-supervised adaptation to refine decision boundaries, and collaborative training on unlabeled target data using confidence-filtered pseudo-labels. 
Comprehensive experiments on multi-site datasets show that CDA generally outperforms state-of-the-art methods in two classification tasks. 
These results highlight its strong generalizability under domain shifts, making it a promising tool for real-world LLD detection.

\if flase
\section*{Declaration of Competing Interest}
The authors declare that they have no known competing financial interests or personal relationships that could have appeared to influence the work reported in this paper.

\section*{CRediT Authorship Contribution Statement}
{\textbf{Y.~Gao}}: Methodology, Software, Writing - original draft. 
{\textbf{Q.~Wang}}:  Writing - review \& editing. 
{\textbf{Y.~Sun}}: Methodology, Writing - review \& editing. 
{\textbf{C.~Wang}}: Writing - review \& editing. 
{\textbf{Y.~Liang}}: Writing - review \& editing. 
{\textbf{M.~Liu}}: Conceptualization, Validation, Writing - review \& editing, Supervision.
\fi 

\section*{Acknowledgments}
The authors gratefully acknowledge Drs. Guy G. Potter, David C. Steffens, and Lihong Wang for their efforts in the data collection of the NBLOD and NCODE cohorts. 
This work was finished when Y. Gao and C. Wang were visiting the University of North Carolina at Chapel Hill. 
Part of the data used in this paper was obtained from the Alzheimer's Disease Neuroimaging Initiative (ADNI) database. 
The investigators within the ADNI contributed to the design and implementation of ADNI and provided data but did not participate in analysis or writing of this article. 
A complete listing of ADNI investigators can be found online (https://adni.loni.usc.edu/wp-content/ uploads/how\_to\_apply/ADNI\_Acknowledgement\_List.pdf).

\footnotesize
\bibliography{refs}

\begin{thebibliography}{10}
\providecommand{\url}[1]{#1}
\csname url@samestyle\endcsname
\providecommand{\newblock}{\relax}
\providecommand{\bibinfo}[2]{#2}
\providecommand{\BIBentrySTDinterwordspacing}{\spaceskip=0pt\relax}
\providecommand{\BIBentryALTinterwordstretchfactor}{4}
\providecommand{\BIBentryALTinterwordspacing}{\spaceskip=\fontdimen2\font plus
\BIBentryALTinterwordstretchfactor\fontdimen3\font minus \fontdimen4\font\relax}
\providecommand{\BIBforeignlanguage}[2]{{%
\expandafter\ifx\csname l@#1\endcsname\relax
\typeout{** WARNING: IEEEtran.bst: No hyphenation pattern has been}%
\typeout{** loaded for the language `#1'. Using the pattern for}%
\typeout{** the default language instead.}%
\else
\language=\csname l@#1\endcsname
\fi
#2}}
\providecommand{\BIBdecl}{\relax}
\BIBdecl

\bibitem{blazer2003depression}
D.~G. Blazer, ``{Depression in late life: Review and commentary},'' \emph{Journal of Gerontology Series A}, vol.~58, no.~3, pp. M249--M265, 2003.

\bibitem{alexopoulos2005vascular}
G.~S. Alexopoulos, ``{The vascular depression hypothesis: 10 years later},'' \emph{Biological Psychiatry}, vol.~60, no.~12, pp. 1304--1305, 2005.

\bibitem{krishnan1997neuroanatomical}
K.~R.~R. Krishnan, W.~M. McDonald, P.~M. Doraiswamy, L.~A. Tupler, M.~Husain, O.~B. Boyko, G.~S. Figiel, and E.~H. Ellinwood~Jr, ``Neuroanatomical substrates of depression in the elderly,'' \emph{Psychological Medicine}, vol.~27, no.~3, pp. 681--692, 1997.

\bibitem{sawyer2012depression}
K.~Sawyer, E.~Corsentino, N.~Sachs-Ericsson, and D.~C. Steffens, ``Depression, hippocampal volume changes, and cognitive decline in a clinical sample of older depressed outpatients and non-depressed controls,'' \emph{Aging \& Mental Health}, vol.~16, no.~6, pp. 753--762, 2012.

\bibitem{mortimer2013neuroimaging}
A.~M. Mortimer, M.~Likeman, and T.~T. Lewis, ``Neuroimaging in dementia: A practical guide,'' \emph{Practical Neurology}, vol.~13, no.~2, pp. 92--103, 2013.

\bibitem{herrmann2008white}
L.~L. Herrmann, M.~Le~Masurier, and K.~P. Ebmeier, ``White matter hyperintensities in late life depression: A systematic review,'' \emph{Journal of Neurology, Neurosurgery \& Psychiatry}, vol.~79, no.~6, pp. 619--624, 2008.

\bibitem{sexton2013systematic}
C.~E. Sexton, C.~E. Mackay, and K.~P. Ebmeier, ``A systematic review and meta-analysis of magnetic resonance imaging studies in late-life depression,'' \emph{The American Journal of Geriatric Psychiatry}, vol.~21, no.~2, pp. 184--195, 2013.

\bibitem{jonsson2019brain}
B.~A. J{\'o}nsson, G.~Bjornsdottir, T.~E. Thorgeirsson, L.~M. Ellingsen, G.~B. Walters, D.~F. Gudbjartsson, H.~Stefansson, K.~Stefansson, and M.~O. Ulfarsson, ``Brain age prediction using deep learning uncovers associated sequence variants,'' \emph{Nature Communications}, vol.~10, no.~1, p. 5409, 2019.

\bibitem{knoll2020deep}
F.~Knoll, K.~Hammernik, C.~Zhang, S.~Moeller, T.~Pock, D.~K. Sodickson, and M.~Akcakaya, ``{Deep-Learning methods for parallel magnetic resonance imaging reconstruction: A survey of the current approaches, trends, and issues},'' \emph{IEEE Signal Processing Magazine}, vol.~37, no.~1, pp. 128--140, 2020.

\bibitem{maartensson2020reliability}
G.~M{\aa}rtensson, D.~Ferreira, T.~Granberg, L.~Cavallin, K.~Oppedal, A.~Padovani, I.~Rektorova, L.~Bonanni, M.~Pardini, M.~G. Kramberger \emph{et~al.}, ``{The reliability of a deep learning model in clinical out-of-distribution MRI data: A multicohort study},'' \emph{Medical Image Analysis}, vol.~66, p. 101714, 2020.

\bibitem{wen2020convolutional}
J.~Wen, E.~Thibeau-Sutre, M.~Diaz-Melo, J.~Samper-Gonz{\'a}lez, A.~Routier, S.~Bottani, D.~Dormont, S.~Durrleman, N.~Burgos, O.~Colliot \emph{et~al.}, ``{Convolutional neural networks for classification of Alzheimer's disease: Overview and reproducible evaluation},'' \emph{Medical Image Analysis}, vol.~63, p. 101694, 2020.

\bibitem{eitel2021promises}
F.~Eitel, M.-A. Schulz, M.~Seiler, H.~Walter, and K.~Ritter, ``Promises and pitfalls of deep neural networks in neuroimaging-based psychiatric research,'' \emph{Experimental Neurology}, vol. 339, p. 113608, 2021.

\bibitem{ganin2015unsupervised}
Y.~Ganin and V.~Lempitsky, ``{Unsupervised domain adaptation by backpropagation},'' in \emph{International Conference on Machine Learning}.\hskip 1em plus 0.5em minus 0.4em\relax PMLR, 2015, pp. 1180--1189.

\bibitem{sarafraz2024domain}
G.~Sarafraz, A.~Behnamnia, M.~Hosseinzadeh, A.~Balapour, A.~Meghrazi, and H.~R. Rabiee, ``{Domain adaptation and gneralization of functional medical data: A systematic survey of brain data},'' \emph{ACM Computing Surveys}, vol.~56, no.~10, pp. 1--39, 2024.

\bibitem{guan2023domainatm}
H.~Guan and M.~Liu, ``{DomainATM: Domain adaptation toolbox for medical data analysis},'' \emph{NeuroImage}, vol. 268, p. 119863, 2023.

\bibitem{long2015learning}
M.~Long, Y.~Cao, J.~Wang, and M.~Jordan, ``{Learning transferable features with deep adaptation networks},'' in \emph{International Conference on Machine Learning}.\hskip 1em plus 0.5em minus 0.4em\relax PMLR, 2015, pp. 97--105.

\bibitem{french2018self}
G.~French, M.~Mackiewicz, and M.~Fisher, ``Self-ensembling for visual domain adaptation,'' in \emph{International Conference on Learning Representations}, no.~6, 2018.

\bibitem{berthelot2019mixmatch}
D.~Berthelot, N.~Carlini, I.~Goodfellow, N.~Papernot, A.~Oliver, and C.~A. Raffel, ``{MixMatch: A holistic approach to semi-supervised learning},'' \emph{Advances in Neural Information Processing Systems}, vol.~32, 2019.

\bibitem{dou2018unsupervised}
Q.~Dou, C.~Ouyang, C.~Chen, H.~Chen, and P.-A. Heng, ``{Unsupervised cross-modality domain adaptation of ConvNets for biomedical image segmentations with adversarial loss},'' in \emph{Proceedings of the 27th International Joint Conference on Artificial Intelligence}, 2018, pp. 691--697.

\bibitem{pan2021disease}
Y.~Pan, M.~Liu, Y.~Xia, and D.~Shen, ``{Disease-image-specific learning for diagnosis-oriented neuroimage synthesis with incomplete multi-modality data},'' \emph{IEEE Transactions on Pattern Analysis and Machine Intelligence}, vol.~44, no.~10, pp. 6839--6853, 2021.

\bibitem{agarwal2021transfer}
D.~Agarwal, G.~Marques, I.~de~la Torre-D{\'\i}ez, M.~A. Franco~Martin, B.~Garc{\'\i}a~Zapira{\'\i}n, and F.~Mart{\'\i}n~Rodr{\'\i}guez, ``{Transfer learning for Alzheimer’s disease through neuroimaging biomarkers: A systematic review},'' \emph{Sensors}, vol.~21, no.~21, p. 7259, 2021.

\bibitem{liu2023transfer}
T.~Liu, Y.~Zhang, Y.~Wu, T.~Li, Z.~Li, W.~Chen, and S.~Zhang, ``{Transfer learning based efficient schizophrenia classification on electroencephalogram (EEG) Signals: A cross-dataset study},'' in \emph{Proceedings of the 2023 15th International Conference on Bioinformatics and Biomedical Technology}, 2023, pp. 143--148.

\bibitem{pei2018multi}
Z.~Pei, Z.~Cao, M.~Long, and J.~Wang, ``{Multi-adversarial domain adaptation},'' in \emph{Proceedings of the AAAI Conference on Artificial Intelligence}, vol.~32, no.~1, 2018.

\bibitem{xu2020adversarial}
M.~Xu, J.~Zhang, B.~Ni, T.~Li, C.~Wang, Q.~Tian, and W.~Zhang, ``{Adversarial domain adaptation with domain mixup},'' in \emph{Proceedings of the AAAI Conference on Artificial Intelligence}, vol.~34, no.~04, 2020, pp. 6502--6509.

\bibitem{zhao2021domain}
A.~Zhao, M.~Ding, Z.~Lu, T.~Xiang, Y.~Niu, J.~Guan, and J.-R. Wen, ``{Domain-adaptive few-shot learning},'' in \emph{Proceedings of the IEEE/CVF Winter Conference on Applications of Computer Vision}, 2021, pp. 1390--1399.

\bibitem{perone2019unsupervised}
C.~S. Perone, P.~Ballester, R.~C. Barros, and J.~Cohen-Adad, ``Unsupervised domain adaptation for medical imaging segmentation with self-ensembling,'' \emph{NeuroImage}, vol. 194, pp. 1--11, 2019.

\bibitem{dosovitskiy2020image}
A.~Dosovitskiy, L.~Beyer, A.~Kolesnikov, D.~Weissenborn, X.~Zhai, T.~Unterthiner, M.~Dehghani, M.~Minderer, G.~Heigold, S.~Gelly \emph{et~al.}, ``{An image is worth 16x16 words: Transformers for image recognition at scale},'' \emph{International Conference on Learning Representations}, 2021.

\bibitem{he2016deep}
K.~He, X.~Zhang, S.~Ren, and J.~Sun, ``{Deep residual learning for image recognition},'' in \emph{Proceedings of the IEEE Conference on Computer Vision and Pattern Recognition}, 2016, pp. 770--778.

\bibitem{butters2008pathways}
M.~A. Butters, J.~B. Young, O.~Lopez, H.~J. Aizenstein, B.~H. Mulsant, C.~F. Reynolds~III, S.~T. DeKosky, and J.~T. Becker, ``Pathways linking late-life depression to persistent cognitive impairment and dementia,'' \emph{Dialogues in Clinical Neuroscience}, vol.~10, no.~3, pp. 345--357, 2008.

\bibitem{mast2008vascular}
B.~T. Mast, T.~Miles, B.~W. Penninx, K.~Yaffe, C.~Rosano, S.~Satterfield, H.~N. Ayonayon, T.~Harris, and E.~M. Simonsick, ``{Vascular disease and future risk of depressive symptomatology in older adults: Findings from the health, aging, and body composition study},'' \emph{Biological Psychiatry}, vol.~64, no.~4, pp. 320--326, 2008.

\bibitem{bora2012gray}
E.~Bora, A.~Fornito, C.~Pantelis, and M.~Y{\"u}cel, ``{Gray matter abnormalities in major depressive disorder: A meta-analysis of voxel based morphometry studies},'' \emph{Journal of Affective Disorders}, vol. 138, no. 1-2, pp. 9--18, 2012.

\bibitem{lai2013gray}
C.-H. Lai, ``{Gray matter volume in major depressive disorder: A meta-analysis of voxel-based morphometry studies},'' \emph{Psychiatry Research: Neuroimaging}, vol. 211, no.~1, pp. 37--46, 2013.

\bibitem{alexopoulos2009serotonin}
G.~S. Alexopoulos, C.~F. Murphy, F.~M. Gunning-Dixon, C.~E. Glatt, V.~Latoussakis, R.~E. Kelly~Jr, D.~Kanellopoulos, S.~Klimstra, K.~O. Lim, R.~C. Young \emph{et~al.}, ``Serotonin transporter polymorphisms, microstructural white matter abnormalities and remission of geriatric depression,'' \emph{Journal of Affective Disorders}, vol. 119, no. 1-3, pp. 132--141, 2009.

\bibitem{gunning2009anterior}
F.~M. Gunning, J.~Cheng, C.~F. Murphy, D.~Kanellopoulos, J.~Acuna, M.~J. Hoptman, S.~Klimstra, S.~Morimoto, J.~Weinberg, and G.~S. Alexopoulos, ``Anterior cingulate cortical volumes and treatment remission of geriatric depression,'' \emph{International Journal of Geriatric Psychiatry}, vol.~24, no.~8, pp. 829--836, 2009.

\bibitem{koenig2014cognition}
A.~M. Koenig and M.~A. Butters, ``{Cognition in late-life depression: treatment considerations},'' \emph{Current Treatment Options in Psychiatry}, vol.~1, no.~1, pp. 1--14, 2014.

\bibitem{benjamin2011structural}
S.~Benjamin and D.~C. Steffens, ``{Structural neuroimaging of geriatric depression},'' \emph{The Psychiatric Clinics of North America}, vol.~34, no.~2, p. 423, 2011.

\bibitem{taylor2013vascular}
W.~D. Taylor, H.~J. Aizenstein, and G.~Alexopoulos, ``{The vascular depression hypothesis: Mechanisms linking vascular disease with depression},'' \emph{Molecular Psychiatry}, vol.~18, no.~9, pp. 963--974, 2013.

\bibitem{kanellopoulos2011hippocampal}
D.~Kanellopoulos, F.~M. Gunning, S.~S. Morimoto, M.~J. Hoptman, C.~F. Murphy, R.~E. Kelly~Jr, C.~Glatt, K.~O. Lim, and G.~S. Alexopoulos, ``{Hippocampal volumes and the brain-derived neurotrophic factor val66met polymorphism in geriatric major depression},'' \emph{The American Journal of Geriatric Psychiatry}, vol.~19, no.~1, pp. 13--22, 2011.

\bibitem{wen2021multidimensional}
J.~Wen, C.~H. Fu, D.~Tosun, Y.~Veturi, Z.~Yang, A.~Abdulkadir, E.~Mamourian, D.~Srinivasan, J.~Bao, G.~Erus \emph{et~al.}, ``{Multidimensional representations in late-life depression: Convergence in neuroimaging, cognition, clinical symptomatology and genetics},'' \emph{arXiv preprint arXiv:2110.11347}, 2021.

\bibitem{tan2023disrupted}
W.~Tan, X.~Ouyang, D.~Huang, Z.~Wu, Z.~Liu, Z.~He, Y.~Long, R.~meta MDD~Consortium \emph{et~al.}, ``{Disrupted intrinsic functional brain network in patients with late-life depression: Evidence from a multi-site dataset},'' \emph{Journal of Affective Disorders}, vol. 323, pp. 631--639, 2023.

\bibitem{ghafoorian2017transfer}
M.~Ghafoorian, A.~Mehrtash, T.~Kapur, N.~Karssemeijer, E.~Marchiori, M.~Pesteie, C.~R. Guttmann, F.-E. De~Leeuw, C.~M. Tempany, B.~Van~Ginneken \emph{et~al.}, ``{Transfer learning for domain adaptation in MRI: Application in brain lesion segmentation},'' in \emph{International Conference on Medical Image Computing and Computer-Assisted Intervention}, 2017, pp. 516--524.

\bibitem{wachinger2021detect}
C.~Wachinger, A.~Rieckmann, and S.~P{\"o}lsterl, ``Detect and correct bias in multi-site neuroimaging datasets,'' \emph{Medical Image Analysis}, vol.~67, p. 101879, 2021.

\bibitem{wolleb2022learn}
J.~Wolleb, R.~Sandk{\"u}hler, F.~Bieder, M.~Barakovic, N.~Hadjikhani, A.~Papadopoulou, {\"O}.~Yaldizli, J.~Kuhle, C.~Granziera, and P.~C. Cattin, ``{Learn to ignore: Domain adaptation for multi-site MRI analysis},'' in \emph{International Conference on Medical Image Computing and Computer-Assisted Intervention}.\hskip 1em plus 0.5em minus 0.4em\relax Springer, 2022, pp. 725--735.

\bibitem{tzeng2017adversarial}
E.~Tzeng, J.~Hoffman, K.~Saenko, and T.~Darrell, ``{Adversarial discriminative domain adaptation},'' in \emph{Proceedings of the IEEE Conference on Computer Vision and Pattern Recognition}, 2017, pp. 7167--7176.

\bibitem{jiang2022disentangled}
K.~Jiang, L.~Quan, and T.~Gong, ``Disentangled representation and cross-modality image translation based unsupervised domain adaptation method for abdominal organ segmentation,'' \emph{International Journal of Computer Assisted Radiology and Surgery}, vol.~17, no.~6, pp. 1101--1113, 2022.

\bibitem{kamnitsas2017unsupervised}
K.~Kamnitsas, C.~Baumgartner, C.~Ledig, V.~Newcombe, J.~Simpson, A.~Kane, D.~Menon, A.~Nori, A.~Criminisi, D.~Rueckert \emph{et~al.}, ``{Unsupervised domain adaptation in brain lesion segmentation with adversarial networks},'' in \emph{International Conference on Information Processing in Medical Imaging}.\hskip 1em plus 0.5em minus 0.4em\relax Springer, 2017, pp. 597--609.

\bibitem{vu2019advent}
T.-H. Vu, H.~Jain, M.~Bucher, M.~Cord, and P.~P{\'e}rez, ``{ADVENT: Adversarial entropy minimization for domain adaptation in semantic segmentation},'' in \emph{Proceedings of the IEEE/CVF Conference on Computer Vision and Pattern Recognition}, 2019, pp. 2517--2526.

\bibitem{guan2021multi}
H.~Guan, Y.~Liu, E.~Yang, P.-T. Yap, D.~Shen, and M.~Liu, ``Multi-site mri harmonization via attention-guided deep domain adaptation for brain disorder identification,'' \emph{Medical Image Analysis}, vol.~71, p. 102076, 2021.

\bibitem{chen2021transunet}
J.~Chen, Y.~Lu, Q.~Yu, X.~Luo, E.~Adeli, Y.~Wang, L.~Lu, A.~L. Yuille, and Y.~Zhou, ``{TransUnet: Transformers make strong encoders for medical image segmentation},'' \emph{arXiv preprint arXiv:2102.04306}, 2021.

\bibitem{hatamizadeh2022unetr}
A.~Hatamizadeh, Y.~Tang, V.~Nath, D.~Yang, A.~Myronenko, B.~Landman, H.~R. Roth, and D.~Xu, ``{UNETR: Transformers for 3D medical image segmentation},'' in \emph{Proceedings of the IEEE/CVF Winter Conference on Applications of Computer Vision}, 2022, pp. 574--584.

\bibitem{liu2021swin}
Z.~Liu, Y.~Lin, Y.~Cao, H.~Hu, Y.~Wei, Z.~Zhang, S.~Lin, and B.~Guo, ``{Swin transformer: Hierarchical vision transformer using shifted windows},'' in \emph{Proceedings of the IEEE/CVF International Conference on Computer Vision}, 2021, pp. 10\,012--10\,022.

\bibitem{huang2020self}
Z.~Huang, H.~Wang, E.~P. Xing, and D.~Huang, ``Self-challenging improves cross-domain generalization,'' in \emph{European Conference on Computer Vision}.\hskip 1em plus 0.5em minus 0.4em\relax Springer, 2020, pp. 124--140.

\bibitem{blum1998combining}
A.~Blum and T.~Mitchell, ``Combining labeled and unlabeled data with co-training,'' in \emph{Annual Conference on Computational Learning Theory}, 1998, pp. 92--100.

\bibitem{shu2018dirt}
R.~Shu, H.~Bui, H.~Narui, and S.~Ermon, ``{A DIRT-T approach to unsupervised domain adaptation},'' in \emph{International Conference on Learning Representations}, 2018.

\bibitem{you2019universal}
K.~You, M.~Long, Z.~Cao, J.~Wang, and M.~I. Jordan, ``Universal domain adaptation,'' in \emph{Proceedings of the IEEE/CVF Conference on Computer Vision and Pattern Recognition}, 2019, pp. 2720--2729.

\bibitem{tang2020unsupervised}
H.~Tang, K.~Chen, and K.~Jia, ``{Unsupervised domain adaptation via structurally regularized deep clustering},'' in \emph{Proceedings of the IEEE/CVF Conference on Computer Vision and Pattern Recognition}, 2020, pp. 8725--8735.

\bibitem{xu2023dual}
R.~Xu, C.~Wang, S.~Xu, W.~Meng, and X.~Zhang, ``{Dual-stream representation fusion learning for accurate medical image segmentation},'' \emph{Engineering Applications of Artificial Intelligence}, vol. 123, p. 106402, 2023.

\bibitem{dong2025multi}
C.~Dong, Y.~Wu, B.~Sun, J.~Bo, Y.~Huang, Y.~Geng, Q.~Zhang, R.~Liu, W.~Guo, X.~Wang \emph{et~al.}, ``A multi-view contrastive learning and semi-supervised self-distillation framework for early recurrence prediction in ovarian cancer,'' \emph{Computerized Medical Imaging and Graphics}, vol. 119, p. 102477, 2025.

\bibitem{ngo2024learning}
B.~H. Ngo, N.-T. Do-Tran, T.-N. Nguyen, H.-G. Jeon, and T.~J. Choi, ``{Learning CNN on ViT: A hybrid model to explicitly class-specific boundaries for domain adaptation},'' in \emph{Proceedings of the IEEE/CVF Conference on Computer Vision and Pattern Recognition}, 2024, pp. 28\,545--28\,554.

\bibitem{steffens2017negative}
D.~C. Steffens, L.~Wang, K.~J. Manning, and G.~D. Pearlson, ``{Negative affectivity, aging, and depression: Results from the neurobiology of late-life depression (NBOLD) study},'' \emph{The American Journal of Geriatric Psychiatry}, vol.~25, no.~10, pp. 1135--1149, 2017.

\bibitem{steffens2004ncode}
D.~C. Steffens, D.~R. McQuoid, and K.~R.~R. Krishnan, ``{The NCODE study: Clinical characteristics and longitudinal outcomes of depressed older adults},'' \emph{International Psychogeriatrics}, vol.~16, pp. 23--35, 2004.

\bibitem{tustison2010n4itk}
N.~J. Tustison, B.~B. Avants, P.~A. Cook \emph{et~al.}, ``{N4ITK: Improved N3 bias correction},'' \emph{IEEE Transactions on Medical Imaging}, vol.~29, no.~6, pp. 1310--1320, 2010.

\bibitem{avants2011ants}
B.~B. Avants, N.~J. Tustison, G.~Song \emph{et~al.}, ``{A reproducible evaluation of ANTs similarity metric performance in brain image registration},'' \emph{NeuroImage}, vol.~54, no.~3, pp. 2033--2044, 2011.

\bibitem{rolls2020aal3}
E.~T. Rolls, C.-C. Huang, C.-K. Lin \emph{et~al.}, ``Automated anatomical labelling atlas 3,'' \emph{NeuroImage}, vol. 206, p. 116189, 2020.

\bibitem{fischl2012freesurfer}
B.~Fischl, ``{FreeSurfer},'' \emph{NeuroImage}, vol.~62, no.~2, pp. 774--781, 2012.

\bibitem{perez-garcia2021torchio}
F.~P{\'e}rez-Garc{\'\i}a, R.~Sparks, and S.~Ourselin, ``{TorchIO: A python library for efficient loading, preprocessing, augmentation and patch-based sampling of medical images in deep learning},'' \emph{Computer Methods and Programs in Biomedicine}, vol. 208, p. 106236, 2021.

\bibitem{cardoso2022monai}
M.~J. Cardoso, W.~Li, R.~Brown, N.~Ma, E.~Kerfoot, Y.~Wang, B.~Murrey, A.~Myronenko, C.~Zhao, D.~Yang \emph{et~al.}, ``{MONAI: An open-source framework for deep learning in healthcare},'' \emph{arXiv preprint arXiv:2211.02701}, 2022.

\bibitem{lin2017focal}
T.-Y. Lin, P.~Goyal, R.~Girshick, K.~He, and P.~Doll{\'a}r, ``{Focal loss for dense object detection},'' in \emph{Proceedings of the IEEE International Conference on Computer Vision}, 2017, pp. 2980--2988.

\bibitem{saito2018maximum}
K.~Saito, K.~Watanabe, Y.~Ushiku, and T.~Harada, ``{Maximum classifier discrepancy for unsupervised domain adaptation},'' in \emph{Proceedings of the IEEE Conference on Computer Vision and Pattern Recognition}, 2018, pp. 3723--3732.

\bibitem{menendez1997jensen}
M.~L. Men{\'e}ndez, J.~Pardo, L.~Pardo, and M.~Pardo, ``{The Jensen-Shannon divergence},'' \emph{Journal of the Franklin Institute}, vol. 334, no.~2, pp. 307--318, 1997.

\bibitem{kullback1951information}
S.~Kullback and R.~A. Leibler, ``{On information and sufficiency},'' \emph{The Annals of Mathematical Statistics}, vol.~22, no.~1, pp. 79--86, 1951.

\bibitem{touvron2021training}
H.~Touvron, M.~Cord, M.~Douze, F.~Massa, A.~Sablayrolles, and H.~J{\'e}gou, ``{Training data-efficient image transformers \& distillation through attention},'' in \emph{International Conference on Machine Learning}.\hskip 1em plus 0.5em minus 0.4em\relax PMLR, 2021, pp. 10\,347--10\,357.

\bibitem{kunanbayev2024training}
K.~Kunanbayev, V.~Shen, and D.-S. Kim, ``{Training ViT with limited data for Alzheimer's disease classification: An empirical study},'' in \emph{International Conference on Medical Image Computing and Computer-Assisted Intervention}.\hskip 1em plus 0.5em minus 0.4em\relax Springer, 2024, pp. 334--343.

\bibitem{jack2008alzheimer}
C.~R. Jack, M.~A. Bernstein, N.~C. Fox, and et~al., ``{The Alzheimer's disease neuroimaging initiative (ADNI): MRI methods},'' \emph{Journal of Magnetic Resonance Imaging}, vol.~27, no.~4, pp. 685--691, 2008.

\bibitem{zhang2025brain}
L.~Zhang, J.~Wu, L.~Wang, L.~Wang, D.~C. Steffens, S.~Qiu, G.~G. Potter, and M.~Liu, ``{Brain anatomy prior modeling to forecast clinical progression of cognitive impairment with structural MRI},'' \emph{Pattern Recognition}, vol. 165, p. 111603, 2025.

\bibitem{bottou2012stochastic}
L.~Bottou, ``{Stochastic gradient descent tricks},'' in \emph{Neural Networks: Tricks of the Trade}.\hskip 1em plus 0.5em minus 0.4em\relax Springer, 2012, pp. 421--436.

\bibitem{aly2005one}
M.~Aly, ``{Survey on multiclass classification methods},'' California Institute of Technology, Tech. Rep., 2005.

\bibitem{pan2010domain}
S.~J. Pan, I.~W. Tsang, J.~T. Kwok, and Q.~Yang, ``{Domain adaptation via transfer component analysis},'' \emph{IEEE Transactions on Neural Networks}, vol.~22, no.~2, pp. 199--210, 2010.

\bibitem{long2013transfer}
M.~Long, J.~Wang, G.~Ding, J.~Sun, and P.~S. Yu, ``{Transfer feature learning with joint distribution adaptation},'' in \emph{Proceedings of the IEEE International Conference on Computer Vision}, 2013, pp. 2200--2207.

\bibitem{ghifary2016scatter}
M.~Ghifary, D.~Balduzzi, W.~B. Kleijn, and M.~Zhang, ``{Scatter component analysis: A unified framework for domain adaptation and domain generalization},'' \emph{IEEE Transactions on Pattern Analysis and Machine Intelligence}, vol.~39, no.~7, pp. 1414--1430, 2016.

\bibitem{sun2016deep}
B.~Sun and K.~Saenko, ``{Deep CORAL: Correlation alignment for deep domain adaptation},'' in \emph{European Conference on Computer Vision}.\hskip 1em plus 0.5em minus 0.4em\relax Springer, 2016, pp. 443--450.

\bibitem{yu2019transfer}
C.~Yu, J.~Wang, Y.~Chen, and M.~Huang, ``{Transfer learning with dynamic adversarial adaptation network},'' in \emph{IEEE International Conference on Data Mining}.\hskip 1em plus 0.5em minus 0.4em\relax IEEE, 2019, pp. 778--786.

\bibitem{cui2020towards}
S.~Cui, S.~Wang, J.~Zhuo, L.~Li, Q.~Huang, and Q.~Tian, ``{Towards discriminability and diversity: Batch nuclear-norm maximization under label insufficient situations},'' in \emph{Proceedings of the IEEE/CVF Conference on Computer Vision and Pattern Recognition}, 2020, pp. 3941--3950.

\bibitem{zhu2020deep}
Y.~Zhu, F.~Zhuang, J.~Wang, G.~Ke, J.~Chen, J.~Bian, H.~Xiong, and Q.~He, ``{Deep subdomain adaptation network for image classification},'' \emph{IEEE Transactions on Neural Networks and Learning Systems}, vol.~32, no.~4, pp. 1713--1722, 2020.

\bibitem{liang2021source}
J.~Liang, D.~Hu, Y.~Wang, R.~He, and J.~Feng, ``{Source data-absent unsupervised domain adaptation through hypothesis transfer and labeling transfer},'' \emph{IEEE Transactions on Pattern Analysis and Machine Intelligence}, vol.~44, no.~11, pp. 8602--8617, 2021.

\end{thebibliography}
\bibliographystyle{IEEEtran}

\end{document}